\documentclass[acmtog]{acmart}%,draft
\usepackage{booktabs} % For formal tables
\usepackage{amsmath}

\usepackage{footnote}
\usepackage{booktabs}
\usepackage{multirow}
\usepackage{paralist}
\usepackage{enumitem}
\usepackage{autobreak}
\usepackage{overpic}
\usepackage{hyperref}
\usepackage{cleveref}

\allowdisplaybreaks

\acmJournal{TOG}
\acmVolume{1}
\acmNumber{1}
\acmArticle{1}
\acmYear{2020}
\acmMonth{8}

\usepackage{mathtools}
\usepackage{amsmath}

\usepackage{amssymb}
\usepackage{amsfonts}
\usepackage{amsopn}
\usepackage{graphicx} % Necessary to use \scalebox
\usepackage{textcomp}
\usepackage{xfrac}
\usepackage{bbm}
\usepackage{overpic}
\usepackage{subfig}
\usepackage{wrapfig}
\usepackage[ruled,vlined,linesnumbered]{algorithm2e}
\usepackage{multirow}

\usepackage{color}
\definecolor{green}{rgb}{0, 0.5, 0}
\definecolor{orange}{rgb}{0.8, 0.6, 0.2}
\definecolor{red}{rgb}{1.0, 0.0, 0.0}
\definecolor{teal}{rgb}{0.0, 0.4, 0.4}
\definecolor{purple}{rgb}{0.65,0,0.65}
\definecolor{saffron}{rgb}{0.95,0.75,0.2}
\definecolor{turquoise}{rgb}{0.0,0.5,0.5}
\definecolor{brown}{rgb}{0.5, 0.16, 0.16}

\usepackage{overpic}
\usepackage{currfile} % \currfiledir

\newlength\savedwidth

\definecolor{lightgray}{rgb}{0.6, 0.6, 0.6}

\newcommand{\Fig}[1]{Figure~\ref{fig:#1}}
\newcommand{\Eq}[1]{Eq.~(\ref{eq:#1})}

\usepackage[normalem]{ulem}

\renewcommand{\paragraph}[1]{\textbf{#1}}

\newcommand{\hidecomment}[1]{}

\newcommand{\bn}{\mathbf{n}}

\newcommand{\bp}{\mathbf{p}}

\DeclareMathOperator*{\argmax}{arg\,max}

\usepackage{xspace}
\newcommand{\nc}[1]{{\color{black}#1}}
\newcommand{\ys}[1]{{\color{black}#1}}

\begin{document}
% Title portion
\title{SymmetryNet: Learning to Predict Reflectional and Rotational Symmetries of 3D Shapes from Single-View RGB-D Images}

% DO NOT ENTER AUTHOR INFORMATION FOR ANONYMOUS TECHNICAL PAPER SUBMISSIONS TO SIGGRAPH 2019!
\author{Yifei Shi}
\affiliation{%
  \institution{National University of Defense Technology}}
%\email{yifei.j.shi@gmail.com}
\author{Junwen Huang}
\affiliation{%
  \institution{National University of Defense Technology}}
\author{Hongjia Zhang}
\affiliation{%
  \institution{National University of Defense Technology}}
\author{Xin Xu}
\affiliation{%
  \institution{National University of Defense Technology}}
\author{Szymon Rusinkiewicz}
\affiliation{%
  \institution{Princeton University}}
\author{Kai Xu}
\affiliation{%
  \institution{National University of Defense Technology}}
\authornote{Corresponding author: Kai Xu (kevin.kai.xu@gmail.com)}
%\email{kevin.kai.xu@gmail.com}

%\affiliation{%
% \institution{Rajiv Gandhi University}
%\email{aprna_patel@rguhs.ac.in}
%\affiliation{%
%  \institution{Tsinghua University}
%  \streetaddress{30 Shuangqing Rd}
%  \country{China}
%}
%\email{chan0345@tsinghua.edu.cn}
%\author{Ting Yan}
%\affiliation{%
%  \institution{Eaton Innovation Center}
%  \city{Prague}
%  \country{Czech Republic}}
%\email{yanting02@gmail.com}
%\author{Tian He}
%\affiliation{%
%  \institution{University of Virginia}
%  \department{School of Engineering}
%  \city{Charlottesville}
%  \state{VA}
%  \postcode{22903}
%  \country{USA}
%}
%\affiliation{%
%  \institution{University of Minnesota}
%\email{tinghe@uva.edu}
%\author{Chengdu Huang}
%\author{John A. Stankovic}
%\author{Tarek F. Abdelzaher}
%\affiliation{%
%  \institution{University of Virginia}
%  \department{School of Engineering}
%  \city{Charlottesville}
%  \state{VA}
%  \postcode{22903}
%  \country{USA}
%}

%\renewcommand\shortauthors{Zhou, G. et al}

\begin{abstract}
%!TEX root = symmetry_prediction.tex

We study the problem of symmetry detection of 3D shapes from single-view RGB-D images, where severely missing data renders geometric detection approach infeasible. We propose an end-to-end deep neural network which is able to predict both reflectional and rotational symmetries of 3D objects present in the input RGB-D image. Directly training a deep model for symmetry prediction, however, can quickly run into the issue of overfitting. We adopt a multi-task learning approach. Aside from symmetry axis prediction, our network is also trained to predict symmetry correspondences. In particular, given the 3D points present in the RGB-D image, our network outputs for each 3D point its symmetric counterpart corresponding to a specific predicted symmetry.
In addition, our network is able to detect for a given shape multiple symmetries of different types.
We also contribute a benchmark of 3D symmetry detection based on single-view RGB-D images. Extensive evaluation on the benchmark demonstrates the strong generalization ability of our method, in terms of high accuracy of both symmetry axis prediction and counterpart estimation. In particular, our method is robust in handling unseen object instances with large variation in shape, multi-symmetry composition, as well as novel object categories. 
\end{abstract}

\if 0


\fi


%!TEX root = ../sceneparse.tex

\begin{teaserfigure}
   \begin{overpic}[width=1.0\textwidth,tics=10]{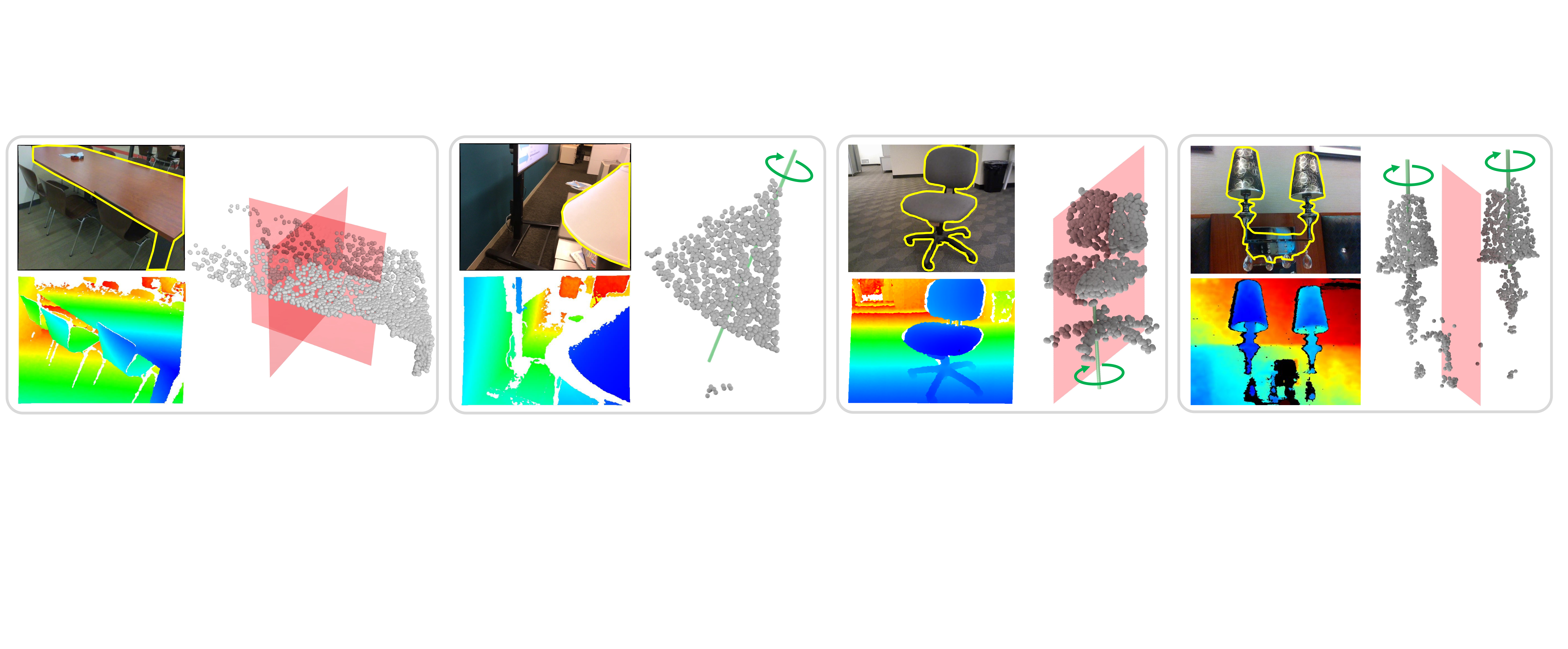}%,grid
   \end{overpic}
   \caption{
   \nc{We propose an end-to-end deep neural network learned to predict both reflectional and rotational symmetries from single-view RGB-D images. For each example, we show the input RGB-D images with the object of interest segmented out (see the yellow masks) as well as the detection results over the unprojected 3D point clouds. Reflectional symmetries are depicted with red planes (reflection plane) and rotational symmetries with green lines (rotation axis). Note how our method is able to detect the composition of an arbitrary number of symmetries, possibly of different types, present in the same object.}
   }
   \label{fig:teaser}
\end{teaserfigure} 

\maketitle

%!TEX root = sceneparse.tex

\section{Introduction}
\label{sec:intro}
Symmetry is omnipresent in both nature and the synthetic world. Symmetry detection is therefore a long-standing problem that has attracted substantial attention in both computer graphics and vision~\cite{liu2010computational,mitra2013symmetry}. Symmetry is at heart a purely geometric concept, with a rigorous definition on the basis of transformation invariance and group theory. It might therefore be supposed that symmetry detection can always be solved by a purely geometric approach. For example, reflectional symmetry in 2D or 3D can be easily parameterized in the transformation space. As such, detection methods such as the Hough transform have historically been utilized to accumulate local cues of symmetry transformations based on detected symmetry point correspondences~\cite{yip2000hough,podolak2006planar,mitra2006partial}.
%As a consequence of this common view, symmetry detection has seldom resorted to learning-based approach.

%There is however one exception, potentially being a practical scenario, where symmetry detection has to rely on learning:
If we consider the problem of symmetry detection in the presence of significant missing data, however, it becomes appropriate to abandon purely-geometric approaches and \emph{infer} what symmetries might be present.  A common application scenario is estimating symmetries of 3D shapes based on a single-view RGB-D image. Single-view symmetry detection finds various potential applications ranging from object/scene completion, camera tracking, and relocalization, to object pose estimation. Due to partial observations and object occlusion, it also poses special challenges that are beyond the reach of geometric detection. For example, it is difficult, if not impossible, to find local symmetry correspondences and transformations supporting global symmetry analysis. In this situation, symmetry analysis should rely not only on geometric detection but also on statistical inference. The latter necessitates data-driven learning.

%It is, however, nontrivial to combine geometric detection and data-driven prediction in one detection framework. The former requires the presence of local support by symmetric point pairs, while the latter needs to ``guess'' the symmetries based on learned shape priors.
In this work, we propose an end-to-end learning approach for symmetry prediction based on a single RGB-D image using deep neural networks. As shown in Figure~\ref{fig:teaser}, given an RGB-D image as input, the network is trained to detect two types of 3D symmetries present in the scene, namely (planar) reflectional and (cylindrical) rotational symmetries, and outputs the corresponding symmetry planes and axes, respectively. Directly training a deep model for symmetry prediction, however, can quickly run into the issue of overfitting. This is due to the fact that the network is able to easily ``memorize'' the symmetry axes of a class of objects in training and will simply perform object recognition at test time. Such an overfitted model cannot generalize well to large shape variation or changes in the symmetries that are present. In fact, symmetry is not a global shape property but rather is supported by local geometric cues: symmetry transformation invariance is defined by local shape correspondences. Straightforward training of symmetry prediction cannot help the network to truly understand the local-to-global support.

%!TEX root = ../sceneparse.tex

\begin{figure*}[t!] \centering
	\begin{overpic}[width=1.0\linewidth,tics=10]{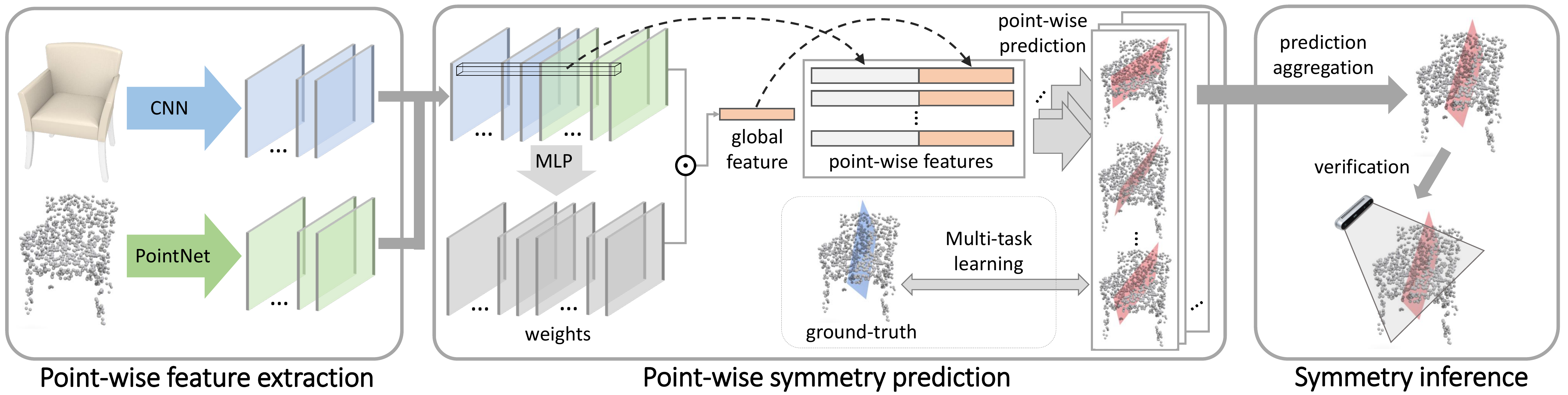}%,grid
   \end{overpic}
   \caption{
   The pipeline of our proposed symmetry prediction method comprising three major components. Taking an RGB image and a depth image as input, the network first extract point-wise appearance and geometry feature maps. The features are then used for point-wise symmetry prediction via multi-task learning. The final symmetry predictions are made with symmetry aggregation and visibility-based verification.
   }
   \label{fig:overview}
\end{figure*} 

To this end, we adopt a multi-task learning approach. Aside from symmetry axis prediction, our network is also trained to predict symmetry correspondences. In particular, given the 3D points present in the RGB-D image, our network outputs for each 3D point its symmetric counterpart corresponding to a specific predicted symmetry. Given any point $P_i$, its counterpart may lie in the 3D point cloud or be missing due to occlusion or limited field of view. To accommodate both cases, we output the counterpart in two modalities. First, we output a heat map for each point in the point cloud indicating the probability of any other points being the symmetric counterpart of this point. Second, we directly regress the $(x,y,z)$ location of this counterpart. To avoid overfitting, we correlate the two tasks by supervising their learning with a unified supervision signal, i.e., the ground-truth local position of the counterparts. We devise a loss that encourages a point with high counterpart probability to be spatially close to its corresponding ground-truth location.

Through proper parameterization, our network is able to handle reflectional symmetry, as well as continuous and discrete rotational symmetry. Since the number of symmetries present in a 3D shape may vary, a network with single output is not suitable for the symmetry prediction task. To this end, we design our network to produce multiple symmetry outputs. When training the network, however, one needs to know how to match the outputs to different ground-truth symmetries in order to compute proper prediction error for gradient propagation. This is achieved by an optimal assignment process, which keeps the entire network end-to-end trainable.

Through extensive evaluation on three symmetry prediction datasets, we demonstrate the strong generalization ability of our method. It attains high accuracy not only for symmetry axis prediction, but also for counterpart estimation. Therefore, our method is robust in handling unseen object instances with large variation in shape, multi-symmetry composition, as well as novel object categories. In summary, we make the following contributions:
\begin{itemize}
  \item We propose the problem of reflective and rotational symmetry detection from single-view RGB-D images, and introduce a robust solution based on deep learning.
  \item We use a series of dedicated tasks (losses) to guide the deep network to learn not only parametrized symmetry axes but also the local symmetry correspondences that support them.
  \item We realize end-to-end learning of multi-symmetry detection by devising an optimal assignment process for multi-output network training.
  \item We propose a benchmark for single-view symmetry detection, encompassing a moderately-sized dataset containing both real and synthetic scenes, as well as evaluation metrics.
\end{itemize}

%!TEX root = sceneparse.tex

\section{Related work}
\label{sec:related}
Symmetry detection has a large body of literature, which has been comprehensively reviewed by the two excellent surveys of Liu et al.~\shortcite{liu2010computational} and Mitra et al.~\shortcite{mitra2013symmetry}. Here we focus only on the work that is most related to our specific designs and techniques.

\paragraph{2D symmetry detection.}
2D symmetry detection has long been a major topic of interest in computer vision. Many approaches have been proposed, including direct methods~\cite{kuehnle1991symmetry}, voting-based methods~\cite{ogawa1991symmetry}, and moment-based methods~\cite{marola1989detection}. Different types of primitive symmetries such as rotation, translation, and reflection, as well as symmetry groups~\cite{liu2000computational}, have been studied. Among all these directions, the detection of bilateral reflectional
symmetry and its skewed version~\cite{liu2001skewed} from 2D images has received the most attention from the community. Our work is relatively closely related to the detection of skewed bilateral symmetry, since the latter is inherently inferring reflectional symmetry of 3D objects from their 2D projections. In contrast to our work, these works do not output symmetries in 3D space and rely on the presence of a large portion of the symmetric regions.

\paragraph{3D symmetry detection.}
Since the two seminal works of Mitra et al.~\shortcite{mitra2006partial} and Podolak et al.~\shortcite{podolak2006planar}, symmetry detection of 3D geometry has attracted much attention in the field of geometry processing. Existing works can be categorized according to different problem settings targeted, such as exact vs.\ approximate symmetry, local vs.\ global symmetry, and extrinsic vs.\ intrinsic symmetry.
Different combinations of the settings lead to different problems and approaches, such as the detection of extrinsic global symmetries~\cite{podolak2006planar,martinet2006accurate}, extrinsic partial symmetries~\cite{mitra2006partial,bokeloh2009symmetry,lipman2010symmetry}, intrinsic global symmetries~\cite{raviv2007,ovsjanikov2008global}, and intrinsic partial symmetries~\cite{xu2009partial,raviv2010full}.
Common to these works is the reliance on 3D shape correspondence~\cite{van2011survey}, which is regarded as a primary building block of 3D symmetry detection. However, in cases of significant missing data, such as single-view scans, shape correspondence becomes extremely challenging.
\ys{Continuous rotational and spherical symmetries can be detected locally by slippage analysis~\cite{gelfand2004shape}. However, it would have to be combined with different methods for other symmetry types.}

\paragraph{Learning-based symmetry detection.}
%Statistical learning-based approaches have been attempted.
Early methods for symmetry detection using statistical learning include the use of feed-forward networks to detect and enhance edges that are symmetric in terms of edge orientation~\cite{zielke1992intensity}. Tsogkas and Kokkinos~\shortcite{tsogkas2012learning} employ hand-crafted features and multiple instance learning to detect ribbon-like structures in natural images, which was later extended to detect more general reflectional symmetry~\cite{shen2016multiple}. Teo et al.~\shortcite{teo2015detection} utilize structured random forests to detect curved reflectional symmetries. Most recently, deep learning has been adopted for the task of 2D symmetry detection~\cite{shen2016object,ke2017srn}, typically detecting reflectional symmetries as 2D skeletons instead of symmetries in 3D space.

Gao et al.~\shortcite{gao2019prs} propose PRS-Net, the first deep learning based symmetry detection method for 3D models demonstrating excellent results. They develop a loss function to measure symmetry correspondence that requires the counterpart of any point to lie on the shape surface. This limits their use in handling single-view scans: the reflective counterpart of a point may be far away from the surface due to missing data, which may lead to high loss and slow convergence.

\paragraph{Learning-based shape correspondence.}
Deep learning has also been applied to shape correspondence.
Existing works mostly focus on learning-based shape descriptors~\cite{huang2017learning}, which have proven more robust than hand-crafted ones. Wei et al.~\shortcite{wei2016dense} learn feature descriptors for each pixel in a depth scan of a human for establishing dense correspondences. Zeng et al.~\shortcite{zeng20173dmatch} learn a local patch descriptor for volumetric data, which can be used for aligning depth data for RGB-D reconstruction. Although data-driven local shape descriptors can be used for symmetry detection, it is unclear how to harness them to realize an end-to-end learned symmetry detector. Moreover, severe data incompleteness renders shape correspondence inapplicable.

\paragraph{Learning-based object pose estimation.}
Our work is also related to 6D object pose estimation based on single-view RGB(D) input, since pose and symmetry usually imply each other. Most existing methods are instance-level~\cite{choi20123d,hodavn2015detection,konishi2018real,georgakis2018matching,avetisyan2019scan2cad,peng2019pvnet,wang2019densefusion} and require a template 3D model, which is unavailable for our single-view symmetry detection task. Recently, Wang et al.~\cite{wang2019normalized} achieve category-level 6D pose estimation based on a Normalized Object Coordinate Space (NOCS), a shared canonical representation of object instances within a category. They train a neural network to directly infer the pixel-wise correspondence between an RGB image and the NOCS. 6D object pose is estimated using shape matching. This method finds difficulty in generalizing across different shape categories. Our network, on the other hand, \ys{attains satisfying cross-category generality on symmetry prediction making it suited for pose estimation.}

%\input{overview}
%!TEX root = sceneparse.tex

\section{Method}
\label{sec:method}

The symmetry of a 3D object is easily measurable when its geometry is fully known. Conventional symmetry detection pipelines for 3D objects normally establish symmetric correspondences within the observed geometrical elements (e.g. points or parts) before aggregating them into meaningful symmetries.
However, single-view observations of real-world objects are usually incomplete due occlusion and limited field of view. Symmetry detection on incomplete geometry is an ill-posed problem which is difficult to solve with existing approaches.

\nc{
When inferring the underlying symmetries of an incompletely observed object, humans usually resolve ambiguities based on whether the object is familiar.
For an object commonly encountered in daily life, a person recognizes its category, estimates its pose, and determines the symmetries, all based on prior knowledge.
For a novel, rarely-encountered object, however, she may look for local evidence of symmetry (i.e., establish symmetry correspondences) over the observed geometry and/or \emph{imagined} unseen parts.
Clearly, symmetry inference for novel objects is much harder since it involves \emph{simultaneous} shape matching and shape completion.
In this work, we propose a unified solution to single-view symmetry detection for both known and novel objects through coupling the predictions of symmetries and symmetry correspondences.
}

Our solution is to train an end-to-end network for symmetry prediction; see \Fig{overview}.
The network consists of three major components. The first module takes an RGB image and a depth image as input and extracts point-wise appearance and geometric features, respectively.
These features are subsequently utilized for point-wise symmetry prediction.
Finally, the third module performs symmetry aggregation and verification during inference.

%In this section, we first define the problem (sec. \ref{sec:definition}), and describe the symmetry prediction network (sec. \ref{sec:network}). We then introduce the inference performed by our method (sec. \ref{sec:inference}).

%!TEX root = sceneparse.tex

\subsection{Problem definition}
\label{sec:definition}

Given an RGB-D image of an 3D object, our goal is to detect its extrinsic reflectional and/or rotational symmetries, if any. In particular, we detect at most $M^\text{ref}$ reflectional symmetries $\mathcal{S}^\text{ref}=\{S^\text{ref}_{i}\}_{i=1,\ldots,M^\text{ref}}$, which is parameterized as $S^\text{ref}_{i}=\{\bp^\text{ref}_{i},\bn^\text{ref}_{i}\}$ with $\bp^\text{ref}_{i}$ being a point in the reflection plane and $\bn^\text{ref}_{i}$ the plane normal.
We also detect at most $M^\text{rot}$ rotational symmetries $\mathcal{S}^\text{rot}=\{S^\text{rot}_{i}\}_{i=1,\ldots,M^\text{rot}}$, parameterized as
$S^\text{rot}_{i}=\{\bp^\text{rot}_{i},\bn^\text{rot}_{i}\}$
where $\bp^\text{rot}$ is a point lying on the rotation axis and $\bn^\text{rot}$ defines the axis orientation.
All symmetries are represented in the camera reference frame.

%!TEX root = sceneparse.tex

\subsection{Symmetry prediction network}
\label{sec:network}
Let us first introduce how the network predicts one symmetry, and then extend it to output multiple symmetries.

\paragraph{Dense-point symmetry prediction.}
We first extract features for both RGB and depth images and then fuse their feature maps.
%There are a variety number of works that could used to this task
Following~\cite{wang2019densefusion}, we extract point-wise appearance and geometric features using a fully-convolutional network~\cite{wang2019densefusion} and a PointNet~\cite{qi2017pointnet}, respectively.
The two features are then concatenated and used for point-wise prediction tasks.
%
%We build our dense-point symmetry prediction network based on the recently proposed feature fusion method for heterogeneous color and 3D point cloud data [pointfusion].
Our network makes individual predictions for each point before aggregating all the predictions to form the final one.
The overall prediction loss is
%\begin{equation}\label{eq:dense_loss}
$\mathcal{L} = \frac{1}{N}\sum_{i}^{N}\mathcal{L}_i$,
%\end{equation}
where $\mathcal{L}_i$ is the prediction loss of point $P_i$.

\ys{Since symmetry is non-local, both local shape properties and the global shape structure are crucial for its detection.
Therefore,} the point-wise prediction takes both point-wise and global features as input.
To compute global features, a straightforward way would be to perform average- or max-pooling over all point features.
However, average-pooling over all points is redundant for symmetry detection, which can be determined by features of sparse points~\cite{mitra2006partial}. On the other hand, max-pooling may lose too much information.
We instead opt for spatially weighted pooling~\cite{hu2017deep}.
This method measures the significance of the each point by learning a weighted mask for every feature map.
We insert a spatially weighted pooling layer after the appearance and geometric feature extraction layers.
The resulting global feature is then concatenated with the point-wise features for symmetry prediction.

%For the feature pooling of point $p_i$, on the neighboring samples in the feature space. The pooled feature contains information from similar points in the symmetry-aware feature space, thus indicating the potential symmetric counterpart implicitly.
%The pooled feature is then concatenated with the point feature and used to predict the symmetries.

%Multi-task learning is prevalent in numerous deep learning applications~\cite{argyriou2007multi}.
To improve the prediction accuracy and generality, we train the point-wise symmetry prediction network with a multi-task learning scheme. In particular, the tasks include 1) a classification of \emph{symmetry type} (\texttt{null} if there is no symmetry), 2) a regression predicting the \emph{symmetry parameters}, %In order to enhance the network generalization ability on untrained objects, we add two extra output channels to
3) a regression estimating \emph{the location of the symmetric counterpart} of a given point for the corresponding symmetry, and 4) a classification indicating \emph{whether an input point is the symmetric counterpart} of the current point.
%a consistence loss to penalize the inconsistency between the predicted symmetry and the counterparts.
To make the point-wise prediction easy to train, all predicted coordinates are represented in a local reference frame centered at the current point, with the same orientation as the camera reference frame.

Although the extra tasks of symmetric counterpart prediction make the point-wise symmetry detection over-constrained, they allow the network to learn the essence of symmetry (i.e., symmetry correspondence)
via reinforcing the relation between symmetry parameters and symmetric counterparts.

%!TEX root = ../sceneparse.tex

\begin{figure}[t!]
	\begin{overpic}[width=1.0\linewidth,tics=10]{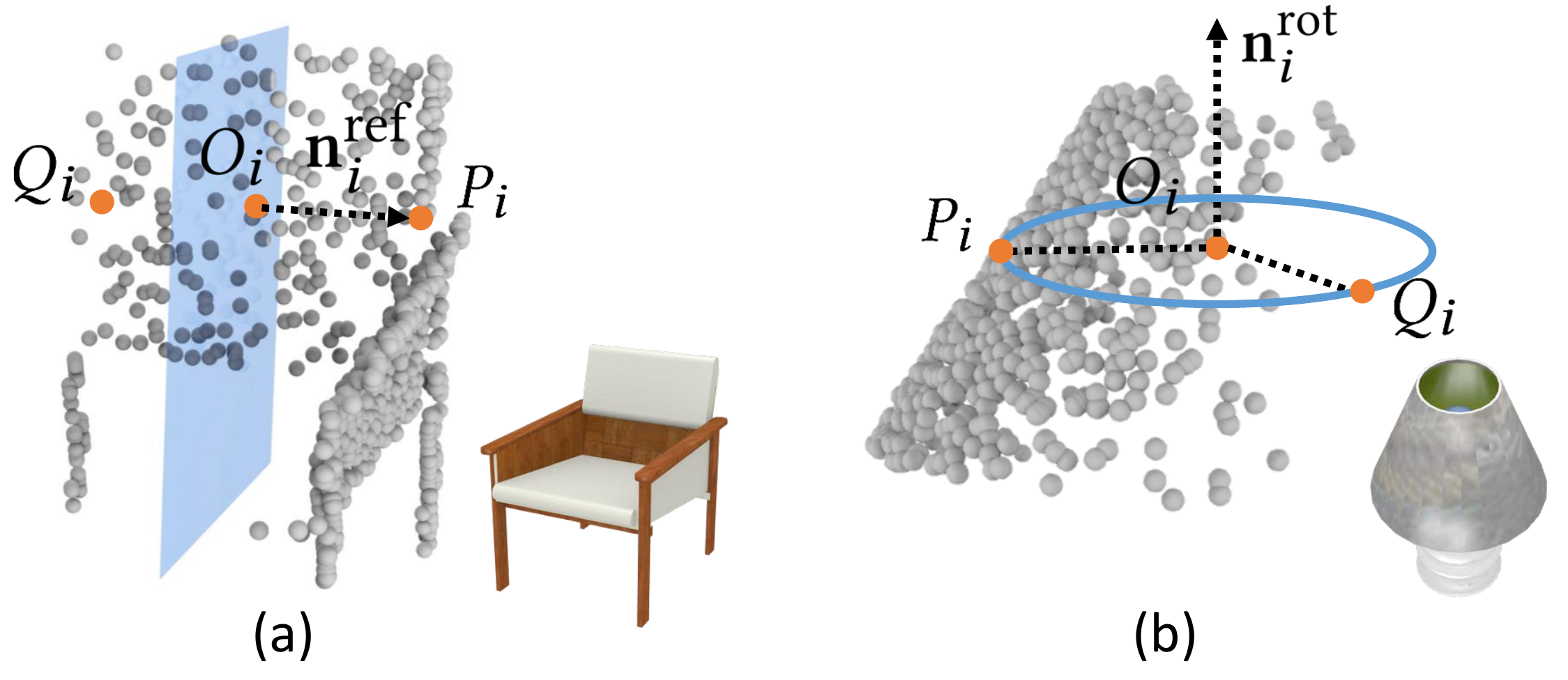}%,grid
   \end{overpic}
   \caption{
   Demonstration of the relationship between the to-be-predicted point $P_i$, its counterpart $Q_i$, and $O_i$ which is the projection of $P_i$ onto (a) the predicted symmetry plane with the plane normal $\bn^\text{ref}_i$ for reflectional symmetry and (b) the predicted symmetry axis with the unit vector $\bn^\text{rot}_i$ for rotational symmetry. 
   }
   \label{fig:loss}
\end{figure} 

Given a point $P_i$, its symmetry prediction loss is defined as
\begin{equation}\label{eq:point_loss}
  \mathcal{L}_i = \mathcal{L}^\text{type}_i + \mathcal{L}^\text{sym}_i,
\end{equation}
where $\mathcal{L}^\text{type}_i$ is the cross-entropy loss for symmetry type classification (\texttt{null} (0) for no symmetry, 1 for reflectional symmetry, 2 for rotational symmetry).
%(0 for no symmetry, 1 for reflectional symmetry and 2 for rotational symmetry).
$\mathcal{L}^\text{sym}_i$ is the loss for symmetry parameters and symmetric counterparts
calculated based on the ground-truth symmetry type:
\begin{equation}\label{eq:symmetry_loss}
\mathcal{L}^\text{sym}_i=
\left\{
    \begin{array}{ll}
    \mathcal{L}^\text{ref\_reg}_i+w^\text{ref}\cdot \mathcal{L}^\text{ref\_cp}_i, & \text{if\ ref.\ sym.}\\
    \mathcal{L}^\text{rot\_reg}_i+w^\text{rot}\cdot \mathcal{L}^\text{rot\_cp}_i, & \text{if\ rot.\ sym.}\\
    0, & \text{if\ no\ sym.}
    \end{array}
\right.
  %L^{sym}_i = L^{cls}_i+L^{reg}_i+L^{cp}_i+L^{cst}_i
\end{equation}

%!TEX root = ../sceneparse.tex

\begin{figure*}[t]
	\begin{overpic}[width=1.0\linewidth,tics=10]{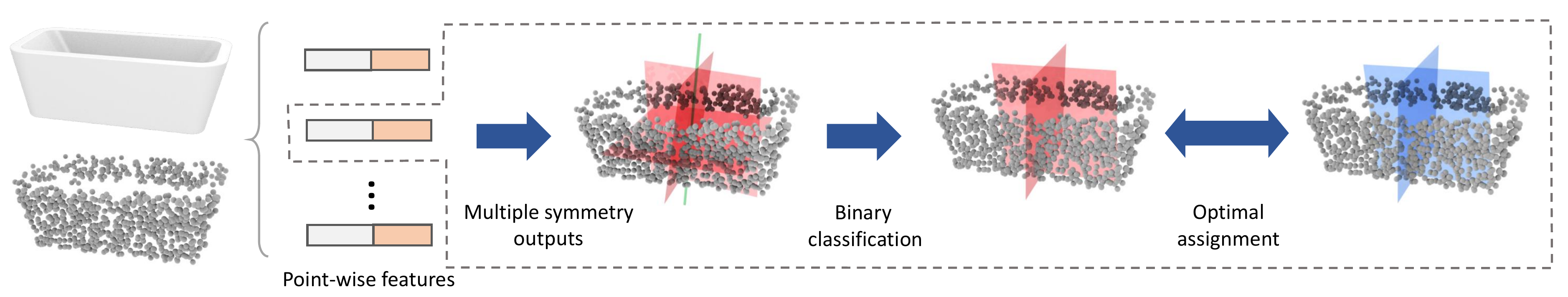}%,grid
   \end{overpic}
   \caption{
   Illustration of the process of obtaining the present symmetries for each point out of multiple output symmetries via binary classification and optimal assignment.
   }
   \label{fig:optimal}
\end{figure*} 

%The network has multiple output channels to represent the symmetry and its counterparts.
%We formulate two types of loss functions for reflectional and rotational symmetries respectively.
For reflectional symmetry (\Fig{loss}a), the network outputs $O_i$, the projection of $P_i$ onto the predicted symmetry plane, and measures the distance between this point and its ground-truth location $\hat{O}_i$:
%, the unit normal of the symmetry plane $\vec{n_i}$
% and the symmetric counterpart $q_i$. The reflectional symmetry prediction loss is:
%\begin{equation}\label{eq:ref_reg_loss}
%    L^{ref\_reg}_i = d_2(t_i,t^*_i)+\left|1-|\vec{n}_i\cdot \vec{n}^*_i|\right|
%\end{equation}
\begin{equation}\label{eq:ref_reg_loss}
    \mathcal{L}^\text{ref\_reg}_i = d^2(O_i, \hat{O}_i),
\end{equation}
where $d$ is Euclidean distance. In the local reference frame of $P_i$, the predicted normal of the symmetry plane $\bn^\text{ref}_i$ is $\frac{\overrightarrow{P_i O_i}}{|\overrightarrow{P_i O_i}|}$.
%, $\vec{n}^*_i$ is the ground-truth unit normal vector of the symmetry plane.

The counterpart loss for reflectional symmetry is:
\begin{equation}\label{eq:ref_ctp_cls_loss}
    \mathcal{L}^\text{ref\_cp}_i = \dfrac{1}{N}\sum_{j}^{N}\mathcal{L}^\text{cls}(p_{ij},\hat{p}_{ij}) + d^2(Q_i,\hat{Q}_i),
\end{equation}
where $p_{ij}$ is the predicted probability that point $P_j$ is the counterpart of point $P_i$ and $\hat{p}_{ij}$ is the ground-truth label (0 for negative and 1 for positive). $\mathcal{L}^\text{cls}$ is the cross-entropy loss. $Q_i$ is the predicted symmetric reflection (counterpart) of $P_i$, and $\hat{Q}_i$ is its corresponding ground-truth.
The counterpart loss penalizes when a point with high counterpart probability is spatially distant from the corresponding ground-truth counterpart location. The weight $w^\text{ref}$ tunes the importance of counterpart prediction.

%The counterpart regression loss for reflectional symmetry is:
%\begin{equation}\label{eq:ref_ctp_reg_loss}
%    L^{ref\_ctp\_reg}_i = d_2(q_i,q^*_i)
%\end{equation}
%where $q^*_i$ is the ground-truth counterpart of point $p_i$.

%The consistence loss for reflectional symmetry is:
%\begin{equation}\label{eq:ref_cst_loss}
%    L^{ref\_cst}_i = \left|2\vec{p_it_i}-\vec{p_iq_i}\right|
%\end{equation}
%this term penalizes when $\vec{p_it_i}$ and $\vec{p_iq_i}$ are not lying on the same line or the length of $\vec{p_it_i}$ is not half of the length of $\vec{p_iq_i}$.

For rotational symmetry (\Fig{loss}(b)), the network predicts point $O_i$ as the projection of $P_i$ onto the predicted symmetry axis $\tilde{\bn}^\text{rot}_i$. We define the rotational symmetry prediction loss as:
\begin{equation}\label{eq:rot_reg_loss}
    \mathcal{L}^\text{rot\_reg}_i = d^2(O_i, \hat{O}_i)+\left|1-|\bn^\text{rot}_i \cdot \hat{\bn}^\text{rot}_i|\right|,
\end{equation}
where $\hat{O}_i$ and $\hat{\bn}^\text{rot}_i$ are the corresponding ground-truths.

Unlike reflectional symmetry, in which each point has only one symmetric counterpart,
rotational symmetry induces more than one counterpart on the rotational orbit.
For discrete rotational symmetry, the number of counterparts is equal to the order of rotational symmetry.
For continuous rotational symmetry, on the other hand, the number is infinite.
Learning to regress all points on the rotation orbit is extremely difficult if not impossible.
We therefore opt to predict the probability for a given input point how likely it is to be in the rotation orbit.
In addition, we predict the order $r$ of the rotational symmetry ($r=0$ for continuous rotational symmetry and $r>0$ for discrete rotational symmetry) using an MLP and a softmax layer for $R$-way classification. $R$ is the maximal order of discrete rotational symmetry in the datatset. We set $R=10$ in our experiment. Note that with this formulation we unify the prediction of continuous and discrete symmetry, leading to reduced model parameters.
%The network predicts a fixed number (C) of counterparts $Q_i = \{q_i^j\},j\in[1,C]$ that are uniformly distributed on the circle.

The counterpart loss for rotational symmetry is:
%\begin{equation}\label{eq:rot_ctp_loss}
%    L^{rot\_ctp}_i = \sum_{q_i^j \in Q_i,q_i^{j*} \in Q^*_i}d_2(q^j_i,q^{j*}_i)
%\end{equation}
\begin{equation}\label{eq:rot_ctp_loss}
    \mathcal{L}^\text{rot\_cp}_i = \dfrac{1}{N}\sum_{j}^{N}\mathcal{L}^\text{cls}(p_{ij}^o,\hat{p}_{ij}^o)+\mathcal{L}^\text{cls}(r_i,\hat{r}_i),
\end{equation}
where $p_{ij}^o$ is the predicted probability that a point $P_j$ lies in the ground-truth orbit of $P_i$,
$r_i$ is the predicted order,
$\hat{p}_{ij}^o$ and $\hat{r}_i$ are the corresponding ground-truths,
and
$w^\text{rot}$ is a trade-off weight.

%Thus, the counterparts of $p_i$ are $Q_i = \{q_i^j\}$ where $q_i^j$ are the points whose probability $prob_i^j$ is larger than a pre-defined threshold (0.8 in our experiments).
%The consistence loss for rotational symmetry is:

%\begin{equation}\label{eq:rot_cst_loss}
%    L^{rot\_cst}_i = \sum_{q_i^j \in Q_i}\left|\left|\vec{p_it_i}\right|-\left|\vec{t_iq_i^j}\right|\right|
%    + \left|\vec{n}_i\cdot \vec{t_iq_i^j}\right|
    %\left|1-\frac{\vec{p_it_i}}{\left|\vec{p_it_i}\right|}\cdot \frac{\vec{p_iq_i}}{\left|\vec{p_iq_i}\right|}\right|
    %+ \left|2-\frac{\left|\vec{p_iq_i}\right|}{\left|\vec{p_it_i}\right|}\right|
%\end{equation}
%where the first term is a penalization when the length of $\vec{p_it_i}$ is not equal to the length of $\vec{t_iq_i^j}$, and the second term acts as a penalization when $\vec{t_iq_i^j}$ is not perpendicular to $\vec{n}$.

%when $\vec{p_it_i}$ and $\vec{p_iq_i}$ not lies in the same line, and the second term penalizes if the length of $\vec{p_it_i}$ is not half of the length of $\vec{p_iq_i}$.

%\begin{equation}\label{eq:symmetry_loss}
%  L_{sym} = Lacxc_{sym\_cls}+L_{sym\_reg}+L_{ctp\_reg}+L_{cst}
%\end{equation}

\nc{
\paragraph{Handling arbitrary number of symmetries.}
To accommodate multiple symmetries in our network,
one option is to train a recurrent neural network which is able to output an arbitrary number of
symmetries sequentially. However, training such a sequential prediction requires a prescribed consistent order for the symmetries, which is obviously infeasible.
A more straightforward option is to have an $M$-way output with $M$ being the maximum number of symmetries per object, which we adopt in our approach. However, training a network with an $M$-way output still requires predefining the order of different outputs. To circumvent this order dependency, we propose an optimal assignment based approach to train the network for order-independent multi-way output.
}

%Even if we restrict our output to 3 reflectional symmetries (as for a cuboid box) and/or a single rotational symmetry (as for a cylinder), using networks with a fixed number of outputs is not feasible. Moreover, the order of reflectional symmetries is not fixed and might not match that of the ground-truth symmetries, making commonly-used multiple output networks (such as Recurrent Neural Networks) inapplicable. As a result, our network should have the ability to automatically decide the number of output symmetries based on the input features.

%We propose to use an optimization scheme during training to achieve this.
In particular, each point produces a maximum of $M^\text{ref}$ outputs for reflectional symmetry or $M^\text{rot}$ for rotational symmetry.
A classifier is used to determine the presence or absence of each symmetry.
For those symmetries verified by the classifier, an optimization is applied to find the maximally-beneficial-matching to the ground-truth symmetries. To be specific, we solve the following optimization during training:
\begin{equation}\label{eq:assignment}
\left.\begin{aligned}
  & \argmax_{\Pi} \sum_{m=1}^{M}\sum_{k=1}^{K}B_{m,k}\Pi_{m,k},\\
  \text{s.t.} \sum_{m=1}^{M}\Pi_{m,k}=1, & \text{ } k\in\{1\ldots K\}; \sum_{k=1}^{K}\Pi_{m,k}\leq 1, m\in\{1\ldots M\}.
\end{aligned}\right.
\end{equation}
$\Pi$ is a permutation matrix with $\Pi_{m,k}\in\{0,1\}$ indicating whether the $k$-th ground-truth symmetry matches the $m$-th predicted symmetries. $M$ is the total number of the output symmetries, and $K$ the total number of the ground-truth symmetries. $B$ is a benefit matrix in which $B_{m,k}$ represents the benefit of matching the $k$-th ground-truth symmetry to the $m$-th predicted symmetry. A higher similarity between two symmetries leads to a larger benefit.

We compute the benefit as follows.
For reflectional symmetries, given two symmetries $S^\text{ref}_{m}$ and $S^\text{ref}_{k}$, and their corresponding reflectional transformations $T^\text{ref}_{m}$ and $T^\text{ref}_{k}$, the benefit of matching $S^\text{ref}_{m}$ and $S^\text{ref}_{k}$ is computed as the Euclidean distance between points that are transformed by the two reflectional transformations respectively:
\begin{equation}\label{eq:benefit_ref}
  B^\text{ref}_{m,k}=\sum_{j}^{N}\frac{1}{\left|T^\text{ref}_{m}(P_j)-T^\text{ref}_{k}(P_j)\right|+\epsilon},
\end{equation}
where $\epsilon=0.01$ is a small value used to prevent dividing by zero.

For rotational symmetries, the benefit of matching two symmetries $S^\text{rot}_{m}$ and $S^\text{rot}_{k}$ is defined as the Euclidean distance between points that are transformed by the two rotational transformations respectively:
\begin{equation}\label{eq:benefit_rot}
  B^\text{rot}_{m,k}=\dfrac{1}{|\Gamma|}\sum_{\gamma \in \Gamma}\sum_{j}^{N}\dfrac{1}{\left|T^{\text{rot},\gamma}_{m}(P_j)-T^{\text{rot},\gamma}_{k}(P_j)\right|+\epsilon},
\end{equation}
where $T^{\text{rot},\gamma}$ is the rotational transformation of $S^\text{rot}$ with a rotation angle of $\gamma$. The set of rotation angles is $\Gamma=\{\kappa\cdot\pi/8\}_{\kappa=1,\ldots,16}$. Note that the transformations with different rotation angles are used only for comparing two rotational symmetries; they have nothing to do with the order of the rotational symmetries.

Solving this optimization amounts to finding the assignment that maximizes the total benefit of matching between the predicted and ground-truth symmetries. We use the Hungarian algorithm~\cite{kuhn1955hungarian} to solve the optimization. Figure~\ref{fig:optimal} shows an illustration of the entire process of outputting multiple symmetries per point and finding the optimal assignment.

%!TEX root = sceneparse.tex

\subsection{Symmetry inference}
\label{sec:inference}

\paragraph{Prediction aggregation.}
During inference, we start by extracting point-wise features and making point-wise symmetry predictions.
We then aggregate these individual predictions to generate the ultimate prediction.
A straightforward method of aggregation is to perform a clustering over the predicted symmetries and select the final predictions as the cluster centers, similar in spirit to~\cite{mitra2006partial} and \cite{podolak2006planar}.
When clustering, we need to account for the importance of the predictions since they are not equally accurate due to the influence of occlusion and non-uniform lighting.
To this end, we introduce a confidence value for each symmetry prediction of each point.
In particular, the confidence is evaluated as the probability output by the softmax layer in predicting symmetry type (the $\mathcal{L}^\text{type}_i$ in \Eq{point_loss}).

After testing the performance of various clustering algorithms, we found that Density-Based Spatial Clustering (DBSCAN)~\cite{ester1996density} works the best for our task. The dissimilarity between two symmetries is defined in \Eq{benefit_ref} and \Eq{benefit_rot}. In addition, we use the confidence value of each predicted symmetry as its density weight, thus encouraging the selection of more confident predictions.

%There are several clustering alternatives that could be used for prediction aggregation. After trying various popular clustering methods, we have chosen DBSCAN since it shows the best performance for our test cases.

%\paragraph{Symmetry refinement}
%Many approaches for point-based tasks use the Iterative Closest Point algorithm (ICP) to perform local refinement. As to our method, an ICP-based refinement process is also beneficial.
%Suppose $T$ is the transformation of the predicted symmetry and $P=\{p_i\} (i\in [1,N])$ is the group of all the points that can be observed. We perform the transformation $T$ on $P$ to obtain the transformed points $T(P)$, followed by using an ICP algorithm to optimize the pose of $T(P)$ w.r.t $P$.
%As a result, the predicted symmetry will be locally refined.
%Note that $P$ only contains the points on the observable surface, so the ICP optimization might not work for all cases.

\paragraph{Visibility-based verification.}
Symmetry prediction on complete 3D objects can be conveniently verified by computing the matching error between the original model and the model transformed by the predicted symmetry. Large matching error implies inaccurate/incorrect symmetry prediction.
This verification, however, is infeasible when the observation is incomplete because even a correct symmetry may have a large matching error due to data incompleteness.
We therefore propose a visibility-based verification approach which is suited to our data, i.e., single-view RGB-D images.

\begin{figure}[t!]
	\begin{overpic}[width=1.0\linewidth,tics=10]{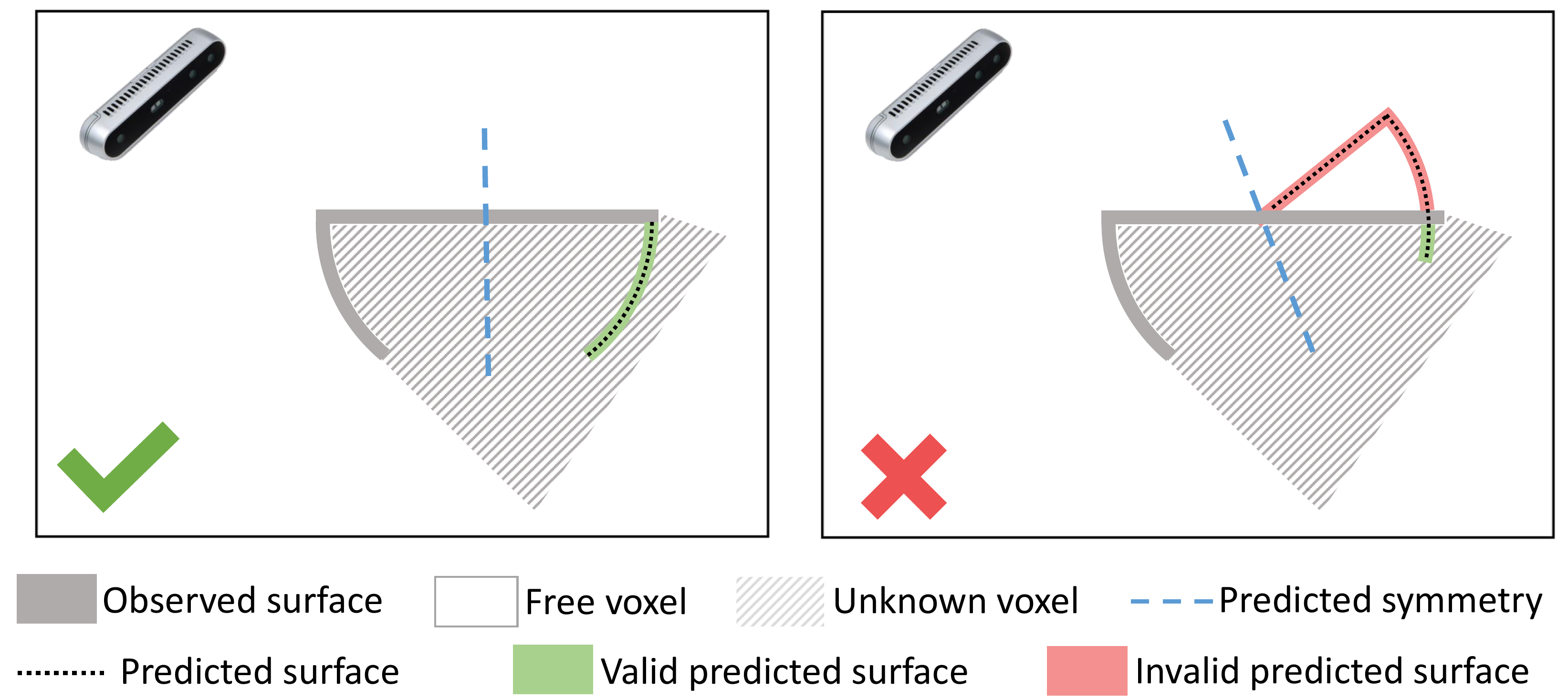}%,grid
   \end{overpic}
   \caption{
   Verification of the symmetry prediction based on the matching error which is equivalent to the overlap between the transformed points $T(P)$ and the regions of space that are known to be free.
   }
   \label{fig:verification}
\end{figure}

As depicted in \Fig{verification}, we first compute a volumetric representation of the space observed by the depth image. Based on visibility w.r.t. the camera pose, the volumetric map contains three types of voxels: observed, free, and unknown. Unknown voxels represent those which are either occluded or outside the FOV. The verification then computes the matching error as the overlap between the transformed surface points and the known free regions. A large overlap means a large confirmed mismatch. We then filter out the predicted symmetries with a large mismatch.

We have also tested using this visibility-based verification as an extra constraint (loss) in the network training. However, we found that it leads to slow convergence while resulting in little improvement in prediction accuracy.

%!TEX root = ../sceneparse.tex

%!TEX root = sceneparse.tex

\section{Implementation details}
\label{sec:impl}

%In this section, we describe the implementation details of the architecture of our network, as well as the training and inference procedures.

\nc{

\paragraph{Network architecture.}
To extract point-wise color features, we use a fully-convolutional network consisting of five convolutional layers, each of which is followed by a Batch Normalization (BN) layer. It encodes an input RGB image of size $H \times W \times 3$ into a $H \times W \times 256$ feature space. We use PointNet to extract geometric features. The architecture of our PointNet implementation is the same as the point cloud segmentation network described in~\cite{qi2017pointnet}. It encodes a point cloud with $N$ points into a $N \times 256$ feature matrix.
The size of the global feature is $512$. After concatenating local and global features, the size of per-point feature is $1024$.
The symmetry predictor is a three-layer Multi-Layer Perceptron (MLP) which takes the per-point features as input and outputs symmetries. Each MLP is followed by a BN layer. The weights $w^\text{ref}$ and $w^\text{rot}$ are both set to $0.5$.
$M^\text{ref}$ and $M^\text{rot}$ in the multi-way prediction of reflectional and rotational symmetries are $9$ and $3$, respectively.

\paragraph{Training and inference.}
We implement the prediction network in PyTorch~\cite{paszke2019pytorch}.
The Adam optimizer~\cite{kingma2014adam} is used with a base learning rate of $0.0001$. We use the default hyper-parameters of $\beta_1 = 0.9$, $\beta_2 = 0.99$, and a weight decay of $0.9$.
The batch size is $32$.
For DBSCAN, the maximum neighborhood distance \texttt{eps} is $0.2$. The minimal number of neighbours for a point to be considered as a core point is $500$.
We filter the symmetry predictions whose confidence value is less than $0.2$.
%We use the implementation of DBSCAN in ?.
For the visibility-based verification, we filter out the predictions with more than $50$ counterpart points located in the known-empty region.

}

%!TEX root = sceneparse.tex

\section{Results and applications}
\label{sec:result}

\subsection{Benchmark}
In order to train and evaluate the proposed network, we have constructed a 3D symmetry detection benchmark for single-view RGB-D images. The benchmark is built upon ShapeNet~\cite{chang2015shapenet}, YCB~\cite{calli2015ycb}, and ScanNet~\cite{dai2017scannet}. For each of the three datasets, we automatically compute symmetries on the 3D models using existing methods. The symmetry labels are then meticulously verified by experienced workers. Finally, we transfer the symmetries of the 3D models to each RGB-D image, transforming by each object's pose. The details of collecting symmetry detection annotations for these datasets are as follows:

\emph{ShapeNet} consists of 3D CAD models with category labels. We first use an optimization-based symmetry detection method to find the ground-truth symmetries in each model, then perform RGB-D virtual scans and transfer the ground-truth symmetry annotations to the local camera coordinates of each RGB-D image. We split this dataset into four subsets: rendered RGB-D images of training models (\emph{Train}), rendered RGB-D images from novel views of training models (\emph{Holdout view}), rendered RGB-D images of testing models (\emph{Holdout instance}), and rendered RGB-D images of models in untrained categories (\emph{Holdout category}). The details of the train and holdout categories are provided in the supplemental material.

\emph{YCB} is a dataset originally built for robotic manipulation and 6D pose estimation.
It contains RGB-D videos of table-top objects with different sizes, shapes, and textures.
High quality 3D reconstructions are provided for each object.
We manually annotate ground-truth symmetries for these reconstructed 3D models, and transfer them to the local camera coordinates of each RGB-D image by using the ground-truth 6D pose of each object.
We follow the original train/test split established in~\citet{calli2015ycb}.

\emph{ScanNet} is a dataset containing RGB-D videos of indoor scenes, annotated with 3D camera poses, surface reconstructions, and semantic segmentations.
The recent work Scan2CAD~\cite{avetisyan2019scan2cad} provides individual alignment between 3D CAD models and the objects present in the reconstructed surfaces.
To obtain the ground-truth symmetries, we first perform an optimization-based symmetry detection on the 3D CAD models, then transfer the detected symmetries to each RGB-D frame.
We split the original train/test split into three subsets: RGB-D images of the training scenes (\emph{Train}), holdout RGB-D images of the training scenes (\emph{Holdout view}), and RGB-D images of the testing scenes (\emph{Holdout scene}).

The statistics of the benchmark datasets are reported in Table \ref{tab:dataset}.

\begin{table}
\caption{
\nc{Statistics of the benchmark.}
}
\label{tab:dataset}
\vspace{-1ex}
\footnotesize\centering
\begin{tabular}{ccrrr}
\toprule
 Dataset & Subset & \#View & \#Object & \#Scene\\
\midrule
\multirow{4}*{\shortstack{ShapeNet}}& Train & $300\,000$ & $30\,000$ & - \\
& Holdout view & $7\,200$ & $2\,400$ & - \\
& Holdout instance & $7\,200$ & $2\,400$ & - \\
& Holdout category & $4\,800$ & $1\,600$ & - \\
\midrule
\multirow{2}*{\shortstack{YCB}}& Train & $16\,189$ & $18$ & $80$ \\
& Test & $2\,949$ & $18$ & $12$ \\
\midrule
\multirow{3}*{\shortstack{ScanNet}}& Train & $13\,126$ & $1\,642$ & $400$ \\
& Holdout view & $4\,723$ & $1\,642$ & $400$ \\
& Holdout scene & $1121$ & $425$ & $100$ \\
\bottomrule
\end{tabular}
\end{table}

\subsection{Evaluation metric}
To evaluate and compare the proposed method, we show precision-recall curves~\cite{funk20172017}
produced by altering the threshold of the confidence value of the prediction.
To determine whether a predicted symmetry is a true positive or a false positive, we compute a dense symmetry error from the difference between the predicted symmetry and the ground-truth symmetry.
Specifically, for a reflectional symmetry, the dense symmetry error of the predicted symmetry $S^\text{ref}$ and the ground-truth symmetry $\hat{S}^\text{ref}$ of an object with points $\mathcal{P}=\{P_i\},i\in [1,N]$ is computed as:
%\begin{equation}\label{eq:metric_ref}
%  \mathcal{E}^\text{ref} = \dfrac{1}{N}\sum_{i}^{N}\dfrac{(T^\text{ref}(P_i)-\hat{T}^\text{ref}(P_i))^2}{\rho},
%\end{equation}
\begin{equation}\label{eq:metric_ref}
  \mathcal{E}^\text{ref} = \dfrac{1}{N}\sum_{i}^{N}\dfrac{\left\|T^\text{ref}(P_i)-\hat{T}^\text{ref}(P_i)\right\|_2}{\rho},
\end{equation}

where $T^\text{ref}$ and $\hat{T}^\text{ref}$ are the symmetric transformations of $S^\text{ref}$ and $\hat{S}^\text{ref}$, respectively, and
$\rho$ is the max distance from the points in $\mathcal{P}$ to the symmetric plane of $\hat{S}^\text{ref}$.

For rotational symmetries, the dense symmetry error between a predicted symmetry $S^\text{rot}$ and the ground-truth symmetry $\hat{S}^\text{rot}$
%of an object with points $P=\{p_i\},i\in [1,N]$ is computed as:
is:

%\begin{equation}\label{eq:metric_rot}
%  \mathcal{E}^\text{rot} = \dfrac{1}{|\Gamma|}\dfrac{1}{N}\sum_{\gamma \in \Gamma}\sum_{i}^{N}\dfrac{(T^{\text{rot},\gamma}(P_i)-\hat{T}^{\text{rot},\gamma}(P_i))^2}{\rho},
%\end{equation}
\begin{equation}\label{eq:metric_rot}
  \mathcal{E}^\text{rot} = \dfrac{1}{|\Gamma|}\dfrac{1}{N}\sum_{\gamma \in \Gamma}\sum_{i}^{N}\dfrac{\left\|T^{\text{rot},\gamma}(P_i)-\hat{T}^{\text{rot},\gamma}(P_i)\right\|_2}{\rho},
\end{equation}
where $T^{\text{rot},\gamma}$ is the rotational transformation of $S^\text{rot}$ with a rotation angle of $\gamma$. The set of rotation angles is $\Gamma=\{\kappa\cdot\pi/8\}_{\kappa=1,\ldots,16}$, and
$\rho$ is the max distance from the points in $\mathcal{P}$ to the rotational axis of $\hat{S}^\text{rot}$.

%where $T^{\text{rot},\gamma}(P_i)$ and $\hat{T}^{\text{rot},\gamma}(P_i)$ are the points of $P_i$ rotated by the symmetric transformation of %$S^\text{rot}$ and $\hat{S}^\text{rot}$, respectively, by angle $\gamma$. $\Gamma=\{\kappa\cdot\pi/8\},\kappa\in[1,16]$.

In all experiments, we set the dense symmetry error threshold to be $0.25$ for both reflectional and rotational symmetries.
\nc{
The confidence value of a predicted symmetry $S$ is computed by the number of input points and the number of samples (symmetries predicted by each point) belonging to the same cluster as symmetry $S$.
}

\subsection{Ablation studies}
To study the importance of each component of our method, we compare our full method against several variants. A specific part of the pipeline is taken out for each variant, as follows:
\begin{compactitem}
\item \textbf{No RGB Input}: without the input RGB image channels (see the first component in Figure~\ref{fig:overview}).  The network can only learn knowledge about symmetries based on geometry.
%\item \textbf{No weighted pooling} Without the weighted pooling layers, the network uses the max pooling layers instead.
\item \textbf{No Counterpart Predictions}: without multi-task learning in the form of counterpart predictions $\mathcal{L}^\text{ref\_cp}$ or $\mathcal{L}^\text{rot\_cp}$ during training (see the second component in Figure~\ref{fig:overview}).
%\item \textbf{No ICP optimization}: No ICP optimization during the inference.
\item \textbf{No Verification}: without the visibility-based filtering of false positives during inference (see the third component in Figure~\ref{fig:overview}).
\end{compactitem}

\begin{figure}[t!]
	\begin{overpic}[width=1.0\linewidth,tics=10]{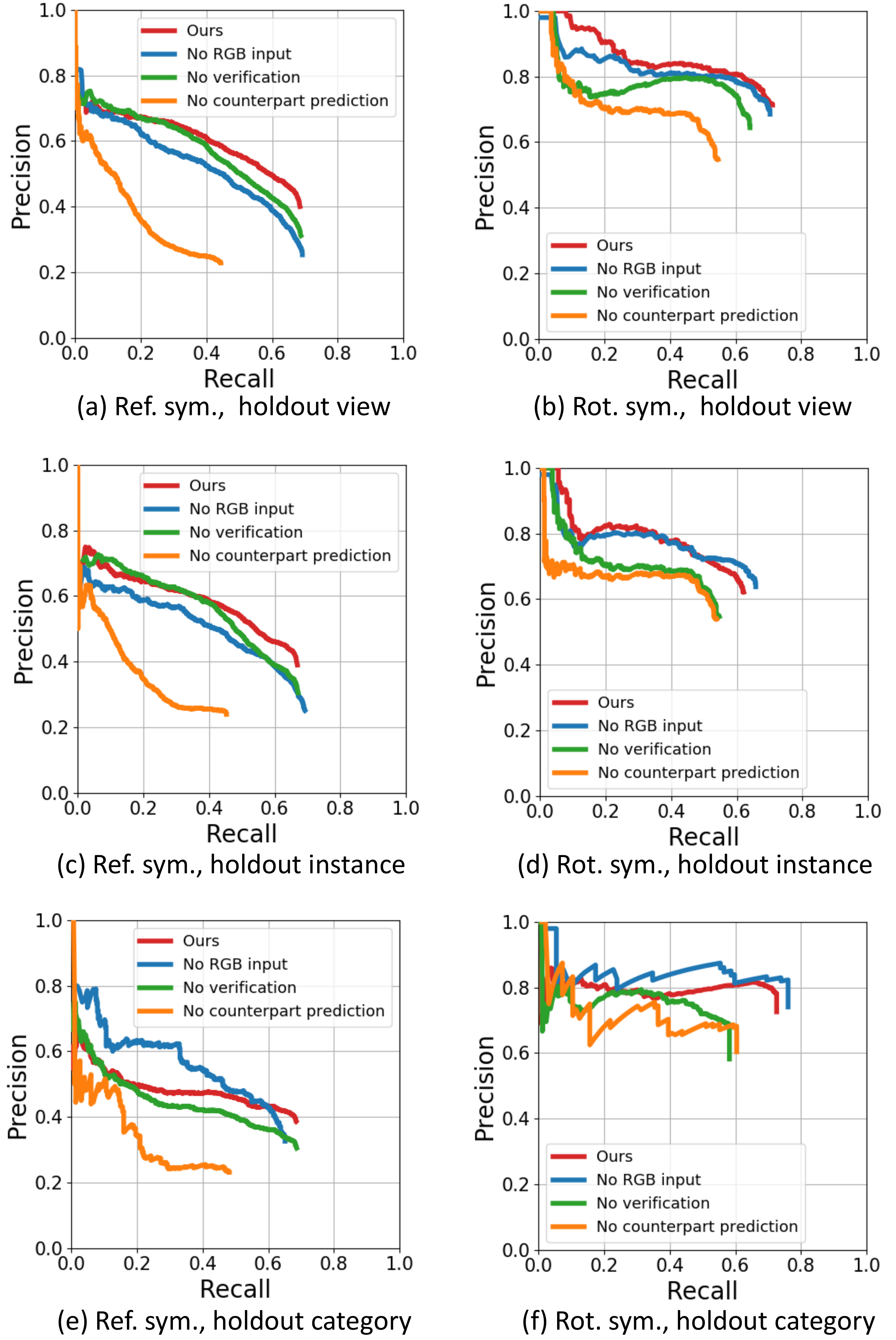}%,grid
   \end{overpic}
   \caption{Ablation studies:
   comparison of symmetry prediction performance on three subsets of ShapeNet between our full proposed method (red) and its several variants (blue: without RGB input; green: without verification procedure; orange: without counterpart prediction, i.e. without multi-task learning). The left and right columns show predictions of reflectional and rotational symmetry, respectively. Note that in most cases using RGB input helps prediction, but the variant without RGB input stands out at predicting the reflectional symmetry for the held-out category (e).
   }
   \label{fig:ablation}
\end{figure}

Figure~\ref{fig:ablation} shows the results of our ablation studies, for reflectional (left column) and rotational (right column) symmetry detection. The full method outperforms the simpler variants in almost all cases.
Omitting counterpart prediction degrades the results the most, especially for reflectional symmetry detection. This demonstrates that the multi-task learning scheme is crucial to our approach.
%No verification baseline is also beneficial to our approach.
An interesting observation is that the baseline without RGB input achieves comparable or even better results on subsets containing novel objects (see Figure~\ref{fig:ablation} d, e, and f). This demonstrates that generalization to unknown objects requires geometry, and confirms the intuition presented in Section~\ref{sec:method}.

\subsection{Comparison to baselines}
We evaluate our method against three baseline symmetry detection methods for objects based on RGB-D images:
\begin{compactitem}
\item \textbf{Geometric Fitting}~\cite{ecins2018seeing}: a state-of-the-art symmetry detection approach for point clouds. It first generates a set of symmetry candidates, and then performs symmetry refinement based on geometric constraints. Since their focus is to detect reflectional symmetries, we only compare with it on the reflectional symmetry prediction task.
%\item \textbf{Transformation space voting}~\cite{mitra2006partial}: a symmetry detection approach for 3D objects that extracts geometric signatures and clusters symmetries using a voting scheme.
%\item \textbf{PRS-Net}: The first learning-based symmetry detection approach for 3D objects with complete geometry. We retrain this network on our dataset.
\item \textbf{RGB-D Retrieval}: an intuitive approach for symmetry detection that finds, for each object in an RGB-D image, the most similar object present in the training data.  The precomputed symmetries are then transferred from the training data to be the symmetry predictions. To achieve this, we train a FoldingNet~\cite{yang2018foldingnet} to extract the feature vectors of all objects in the training data. During testing, $L2$ distance in the feature space is used to retrieve the most similar RGB-D image.
\nc{\item \textbf{Shape Completion}: a two-step approach which first performs a shape completion~\cite{liu2019morphing} on the input point cloud and then detects symmetries on the completed shape by a geometric symmetry detection method~\cite{li2014efficient}. We compare to it on the reflectional symmetry prediction task of ShapeNet.}
\item \textbf{DenseFusion~\cite{wang2019densefusion}}: a cutting-edge approach to estimate the 6D pose of the known objects. We transform the ground-truth symmetries of the known objects by using the predicted 6D pose produced by DenseFusion, thus obtaining the predicted symmetries for each object in the RGB-D images. Since DenseFusion only works on scenarios where the geometries of the target objects are known, we compare to it on the YCB dataset.
\end{compactitem}

%!TEX root = ../sceneparse.tex

\begin{figure}[t!]
	\begin{overpic}[width=1.0\linewidth,tics=10]{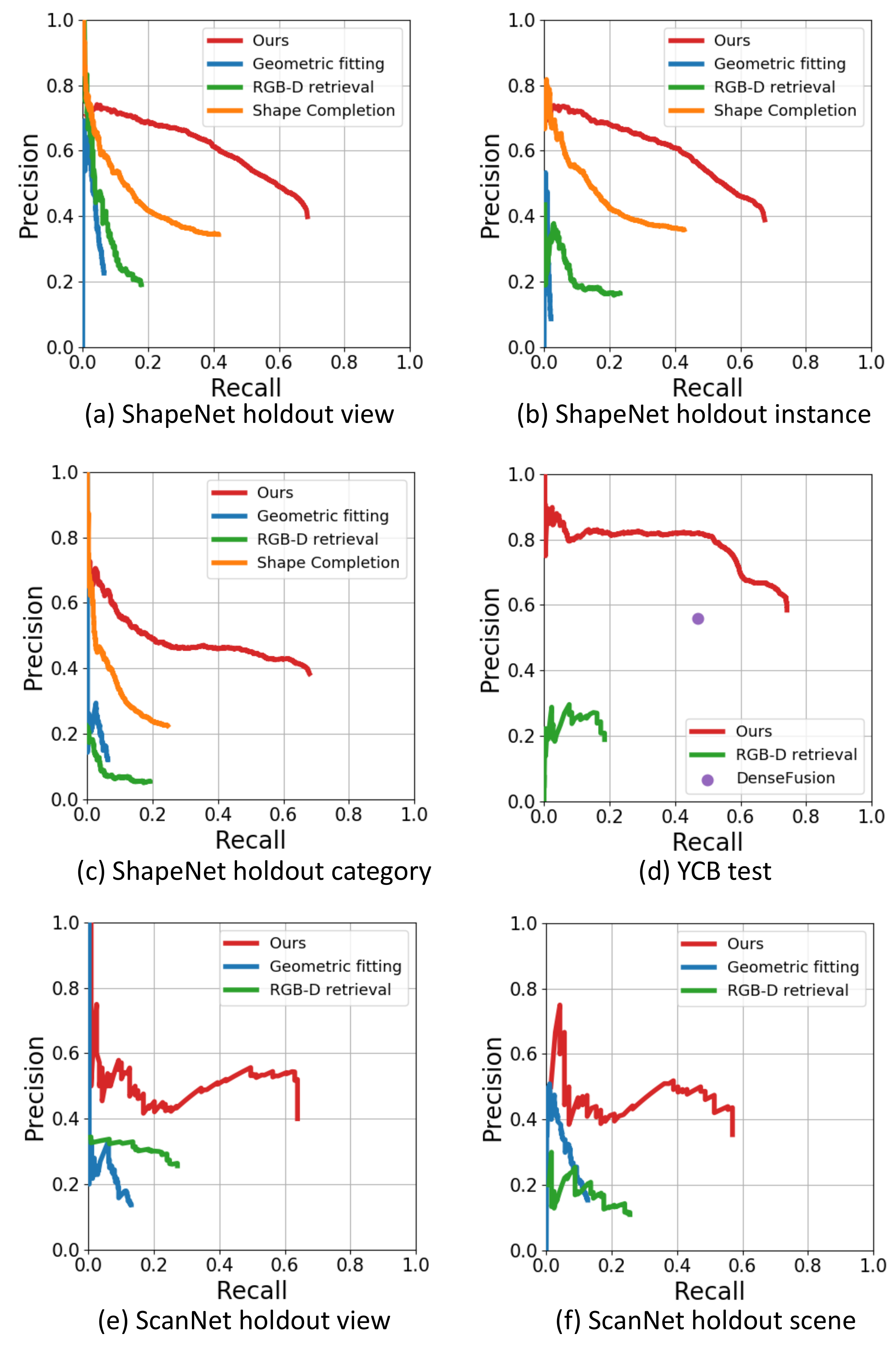}%,grid
   \end{overpic}
   \caption{
   \nc{Comparison between our proposed approach and the baseline methods on the performance of predicting reflectional symmetry for three datasets divided into six subsets: (a) ShapeNet holdout view; (b) ShapeNet holdout instance; (c) ShapeNet holdout category; (d) YCB test; (e) ScanNet holdout view; and (f) ScanNet holdout scene. Note that in (c), our approach is significantly more precise at symmetry prediction for categories that it has never encountered before (holdout category), demonstrating high generalization ability.}
   }
   \label{fig:comp_ref}
\end{figure}

\begin{figure}[t!]
	\begin{overpic}[width=1.0\linewidth,tics=10]{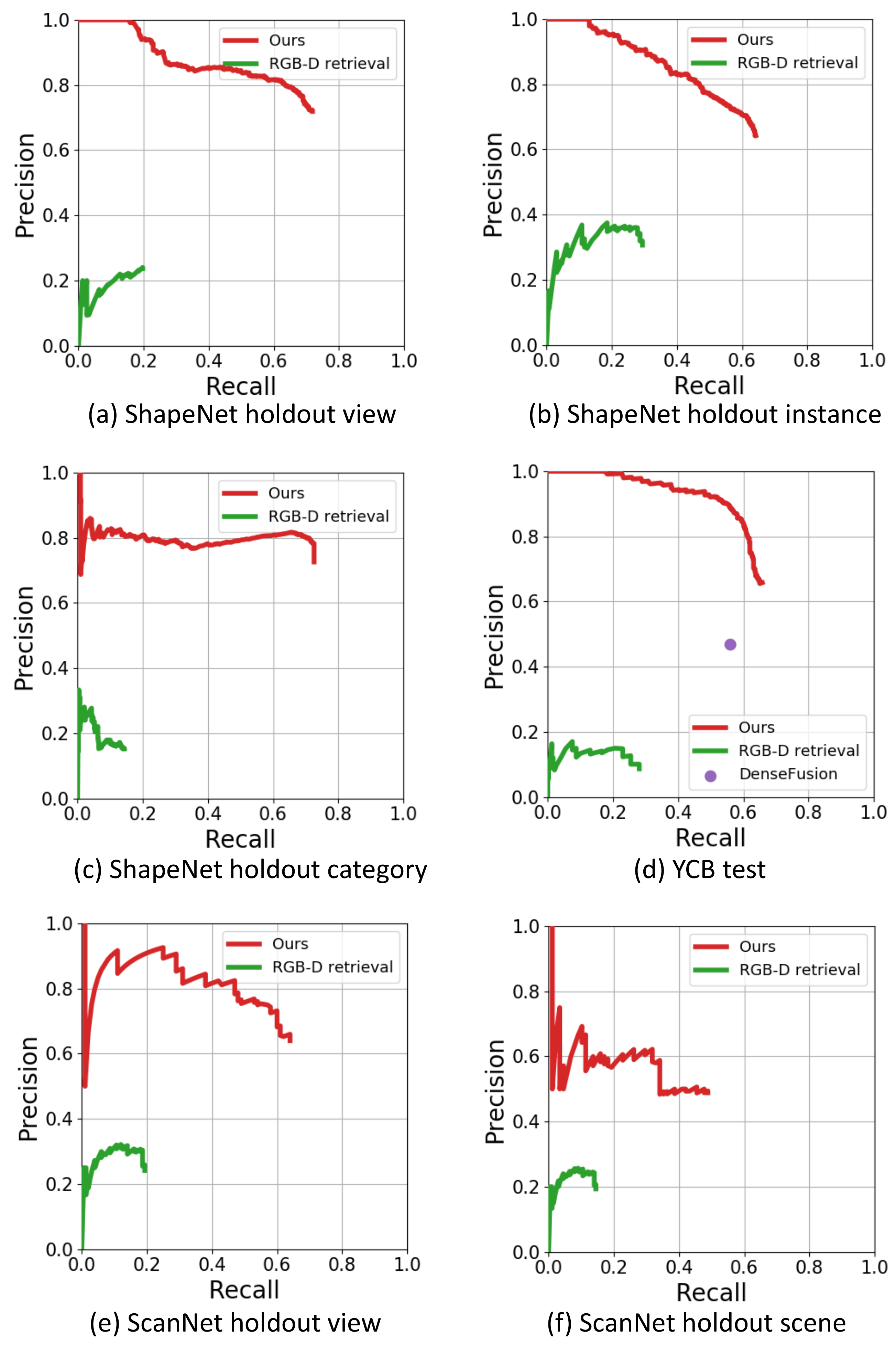}%,grid
   \end{overpic}
   \caption{
   \nc{Comparison between our proposed approach and the baseline methods on the performance of predicting rotational symmetry for three datasets divided into six subsets: (a) ShapeNet holdout view; (b) ShapeNet holdout instance; (c) ShapeNet holdout category; (d) YCB test; (e) ScanNet holdout view; and (f) ScanNet holdout scene. Similar to Figure~\ref{fig:comp_ref}(c), the generalization ability of our approach is illustrated in (c).}
   }
   \label{fig:comp_rot}
\end{figure}

The comparisons are plotted in Figure~\ref{fig:comp_ref} (reflectional symmetry) and Figure~\ref{fig:comp_rot} (rotational symmetry). They show that our method outperforms the baselines by a large margin, for both the reflectional and rotational symmetry detection tasks, and over all the data subsets.
Crucially, our method achieves relatively high performance on subsets (ShapeNet \emph{holdout category} and ScanNet \emph{holdout scene}) that include very different objects from those present in the training data.
\nc{The Shape Completion baseline is inferior to our method, especially on the ShapeNet \emph{holdout Category} subset, due to its poor generality \ys{in completing the shape} on the untrained categories.}
Even though the Geometric Fitting baseline has high precision for the symmetries it detects, it fails to detect most of the symmetries since it can be easily influenced by incomplete observations.
The RGB-D Retrieval baseline shows worse performance than our method, especially on the ShapeNet \emph{holdout Category} subset and the ScanNet dataset, due to its weaker generalization ability.
The DenseFusion baseline achieves a relatively high precision and recall on the YCB dataset. However, it cannot be extended to datasets containing objects that have not been seen during training.

%!TEX root = ../sceneparse.tex

%!TEX root = ../sceneparse.tex

%!TEX root = ../sceneparse.tex

\begin{figure*}[t!] \centering
	\begin{overpic}[width=1.0\linewidth,tics=10]{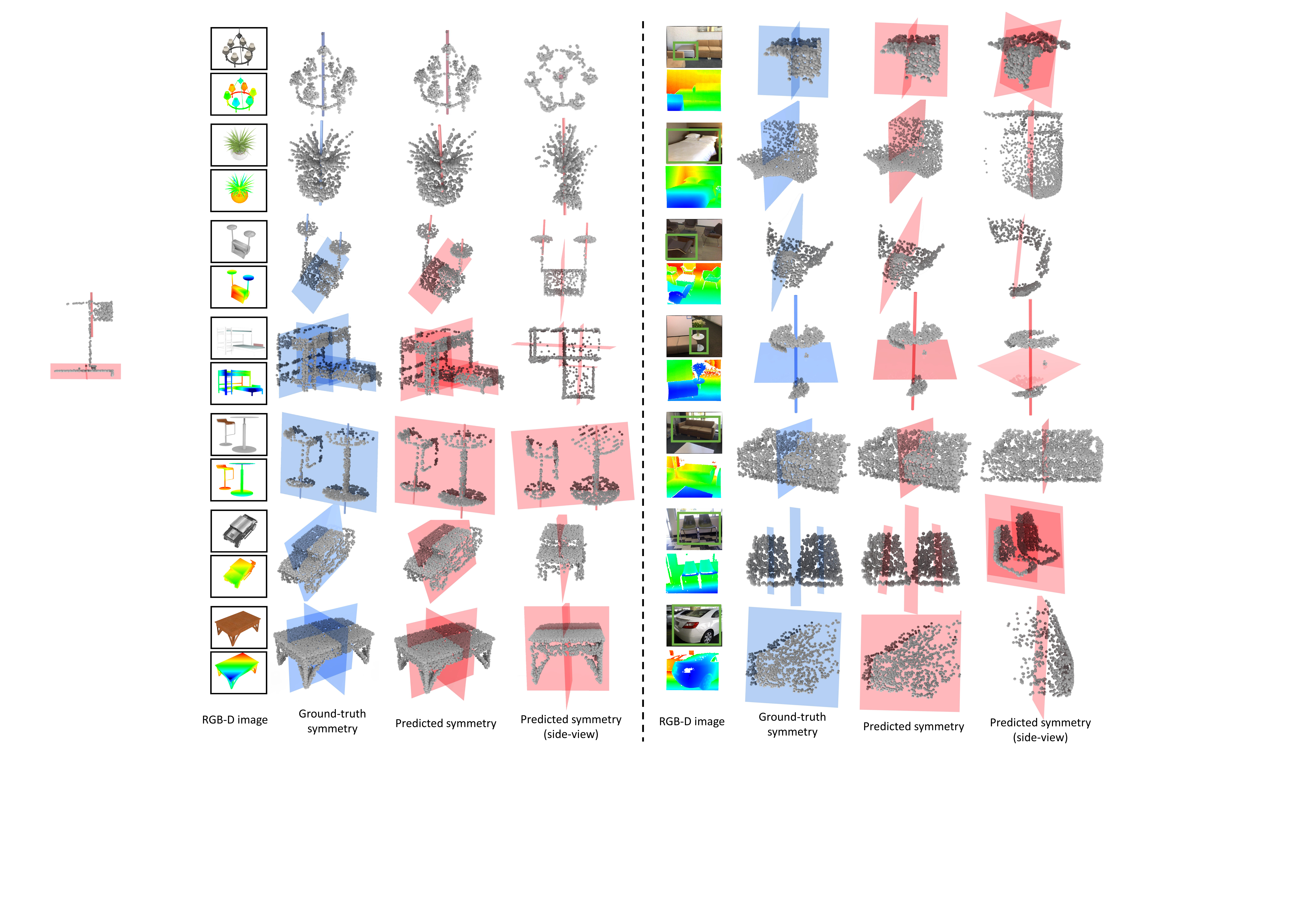}%,grid
   \end{overpic}
   \caption{\nc{Qualitative symmetry prediction results on ScanNet~\cite{dai2017scannet} and ShapeNet~\cite{chang2015shapenet}. The first column has two images of the object, i.e. the input RGB image with the target object marked in green rectangle (upper) and the input depth image (lower). The second column shows the ground-truth symmetries of the objects. The last two columns visualize the predicted symmetries by our method. Our method is able to handle objects with all forms of symmetry compositions (reflectional symmetry only, rotational symmetry only and multi-symmetry).}
   }
   \label{fig:gallery}
\end{figure*}

\subsection{Qualitative results}
\nc{
Figure~\ref{fig:gallery} visualizes the symmetry prediction results for both synthetic and real data.
Our approach detects both reflectional and rotational symmetries in challenging cases, such as novel objects, objects with multiple symmetries, and objects with heavy occlusion.
We also show the qualitative comparison with baselines in Figure~\ref{fig:quan_comp}.
We see that partial observations and occlusion interfere with the ability of Geometric Fitting and PRS-Net to establish correspondences on the observed points.
RGB-D Retrieval provides poor features for untrained objects,
leading to an inability to predict accurate symmetries.
Shape Completion can generate reasonably good geometry for common objects, but it is less capable in cases where the objects are novel or heavily occluded.
Our method, in contrast, successfully predicts symmetries with high accuracy for all of these examples.
}

%\kxc{Add discussion to both figures here.}

%\input{figures/occ_rot}

%!TEX root = ../sceneparse.tex

\begin{figure}[t!]
	\begin{overpic}[width=1.0\linewidth,tics=10]{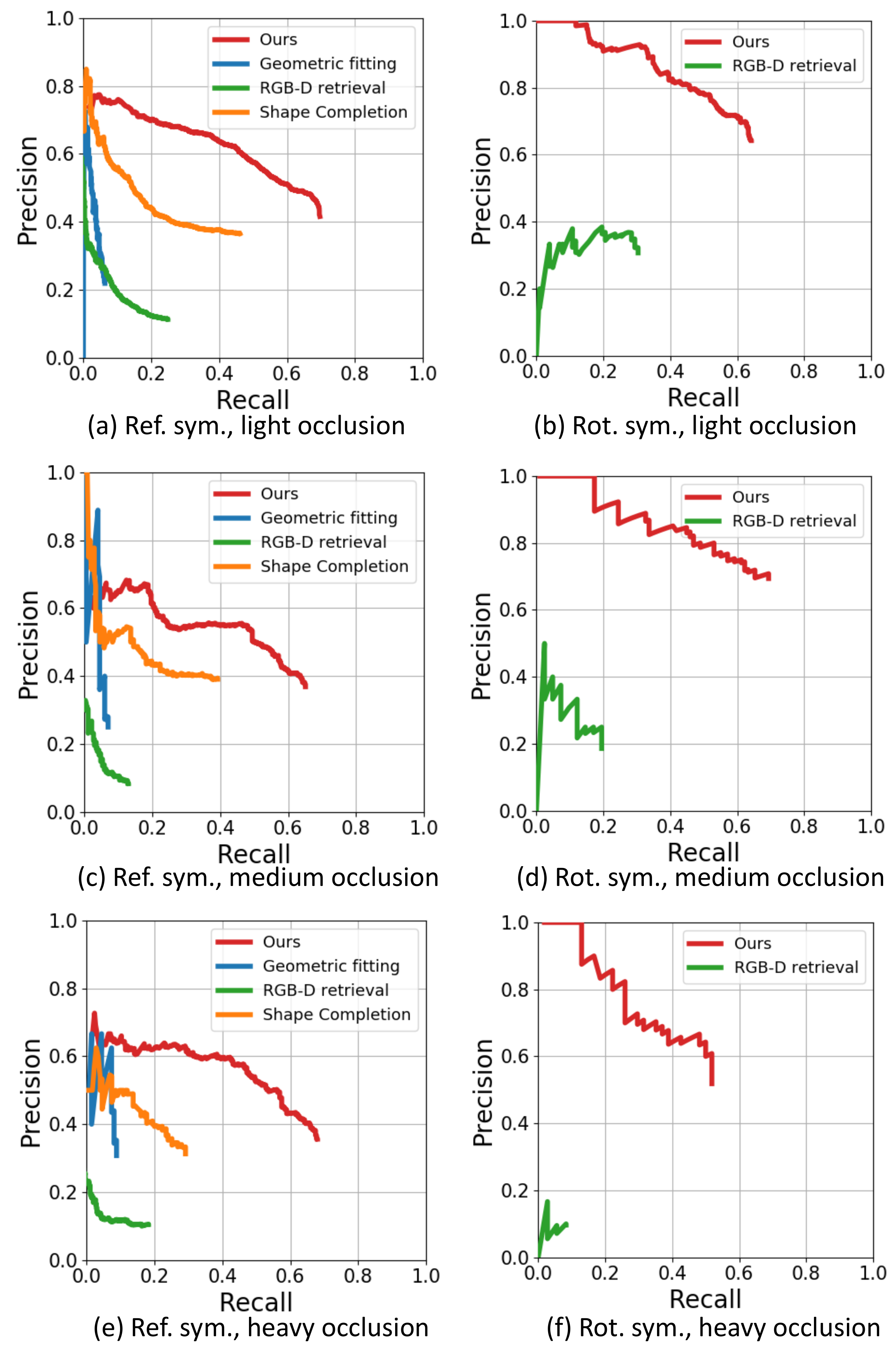}%,grid
   \end{overpic}
   \caption{
   Occlusion ratio sensitivity evaluation of our proposed approach (red), compared to the Geometric Fitting method (blue) and RGB-D Retrieval (green).  The three rows include different occlusion ratios: light (50-60\%) occlusion for (a) and (b); medium (60-70\%) occlusion for (c) and (d); and heavy (70-80\%) occlusion for (e) and (f). The left and right columns illustrate reflectional and rotational symmetry detection, respectively. While the performance of all methods decreases with increasing occlusion, our approach outperforms the baselines in all scenarios, and demonstrates relatively modest decrease in performance with increasing occlusion.
   }
   \label{fig:occ}
\end{figure} 
\subsection{Sensitivity to occlusion}
%\kxc{Show one figure for visual results.}
To evaluate the capability of our approach when it comes to occluded objects, we create a dataset based on ShapeNet by grouping the data according to the occlusion ratio. In order to generate data with mutual occlusion, we randomly add a foreground mask in the rendered RGB-D images.
The occlusion ratio of the object is computed by dividing the area of the occluded region by the area of the whole surface.
Examples of the occluded data are provided in the supplemental material.
Note that both self-occlusion and mutual-occlusion are included.

\begin{figure*}[t!] \centering
	\begin{overpic}[width=1.0\linewidth,tics=10]{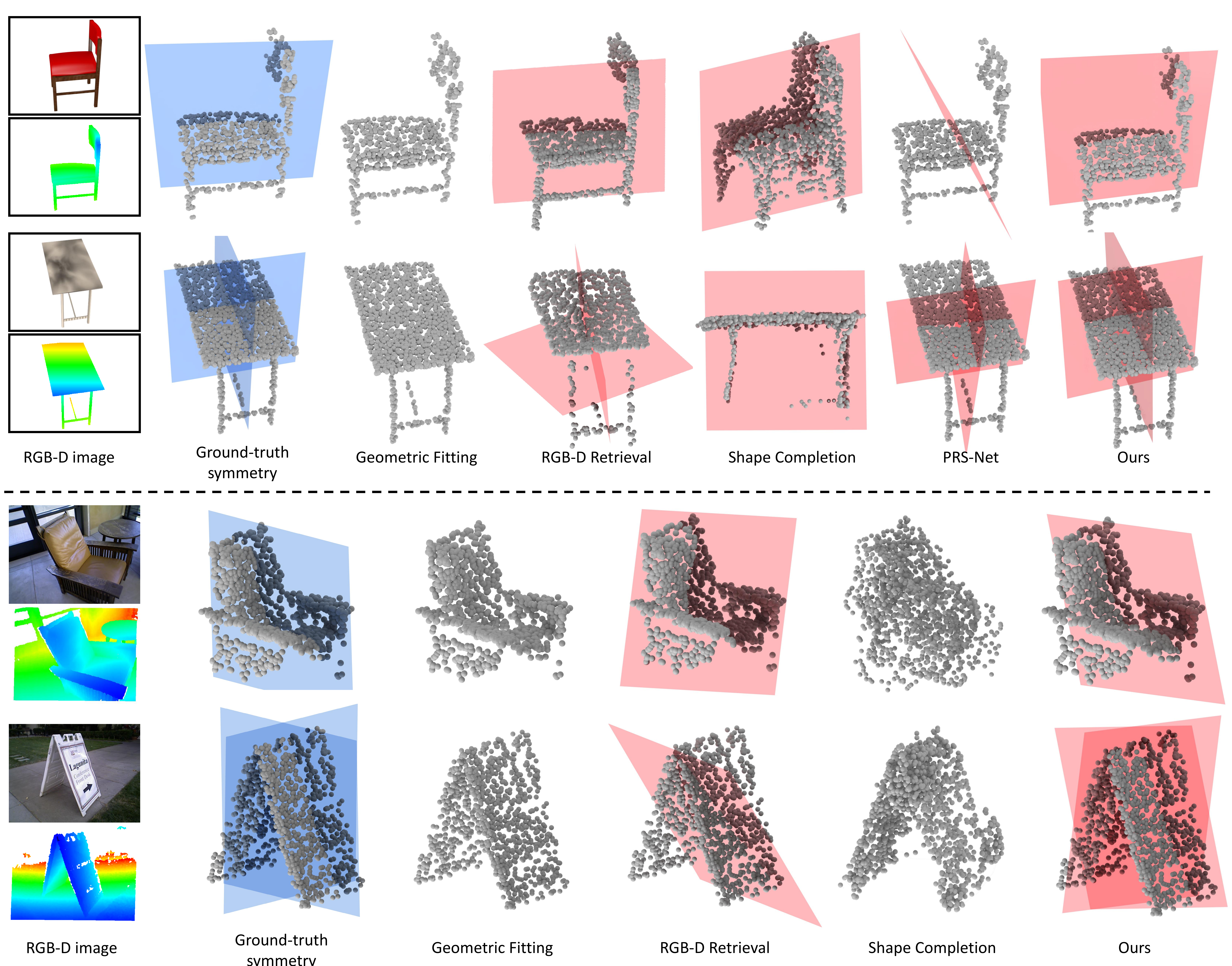}%,grid
   \end{overpic}
   \caption{
   \nc{Qualitative comparisons to previous works (Geometric Fitting~\cite{ecins2018seeing}, RGB-D Retrieval, PRS-Net~\cite{gao2019prs} and Shape Completion~\cite{liu2019morphing,li2014efficient}.) on both synthetic data and real data. The Geometric Fitting baseline fails to detect any symmetry. The PRS-Net baseline correctly predicts the number of symmetries but fails to regress the parameters of the symmetries accurately. The RGB-D Retrieval baseline could not predict the parameters of symmetries correctly. The Shape Completion baseline predicts accurate symmetries for objects with simple geometry, but fails on cases where the objects are novel or occluded. Our method achieves the best performance on the four examples.}
   }
   \label{fig:quan_comp}
\end{figure*}

\nc{
Figure~\ref{fig:occ} compares our approach with the Shape Completion, Geometric Fitting, and RGB-D Retrieval baselines, all three of which are capable of finding symmetries for occluded objects.
It is evident that our method performs better than the baseline methods for all the experiments. While the overall performance is generally affected as the occlusion ratio increases, ours outperforms the baseline methods and shows a relatively smaller decreasing rate in all cases.
Qualitative results on the real data with occlusions are shown in Figure~\ref{fig:teaser}, Figure~\ref{fig:gallery} and Figure~\ref{fig:quan_comp}.
}

\begin{figure}[t!]
	\begin{overpic}[width=1.0\linewidth,tics=10]{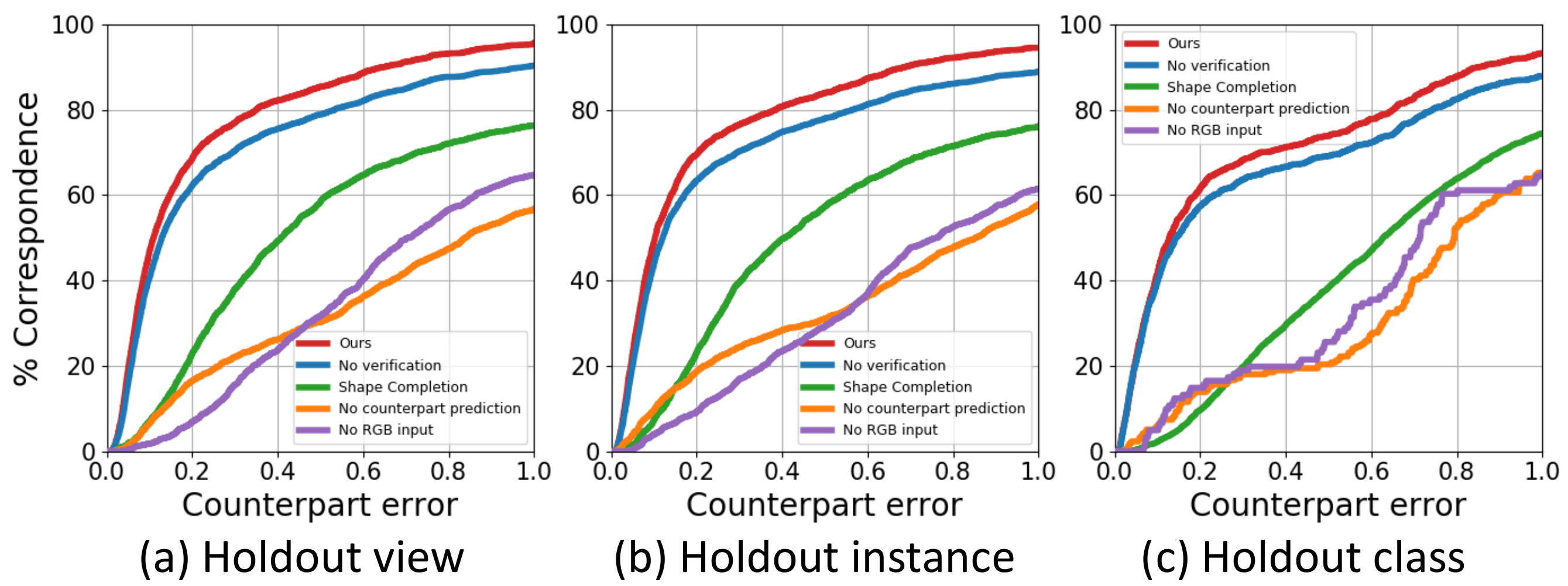}%,grid
   \end{overpic}
   \caption{\nc{Comparison between our proposed approach and the baseline methods on the performance of predicting counterparts \ys{of the reflectional symmetries} on ShapeNet dataset. Each curve depicts the percentage of counterpart correspondences (y-axis) whose Euclidean distance (error) are within the thresholds (x-axis).}
   }
   \label{fig:cpt_eval}
\end{figure}

\begin{figure}[t!]
	\begin{overpic}[width=1.0\linewidth,tics=10]{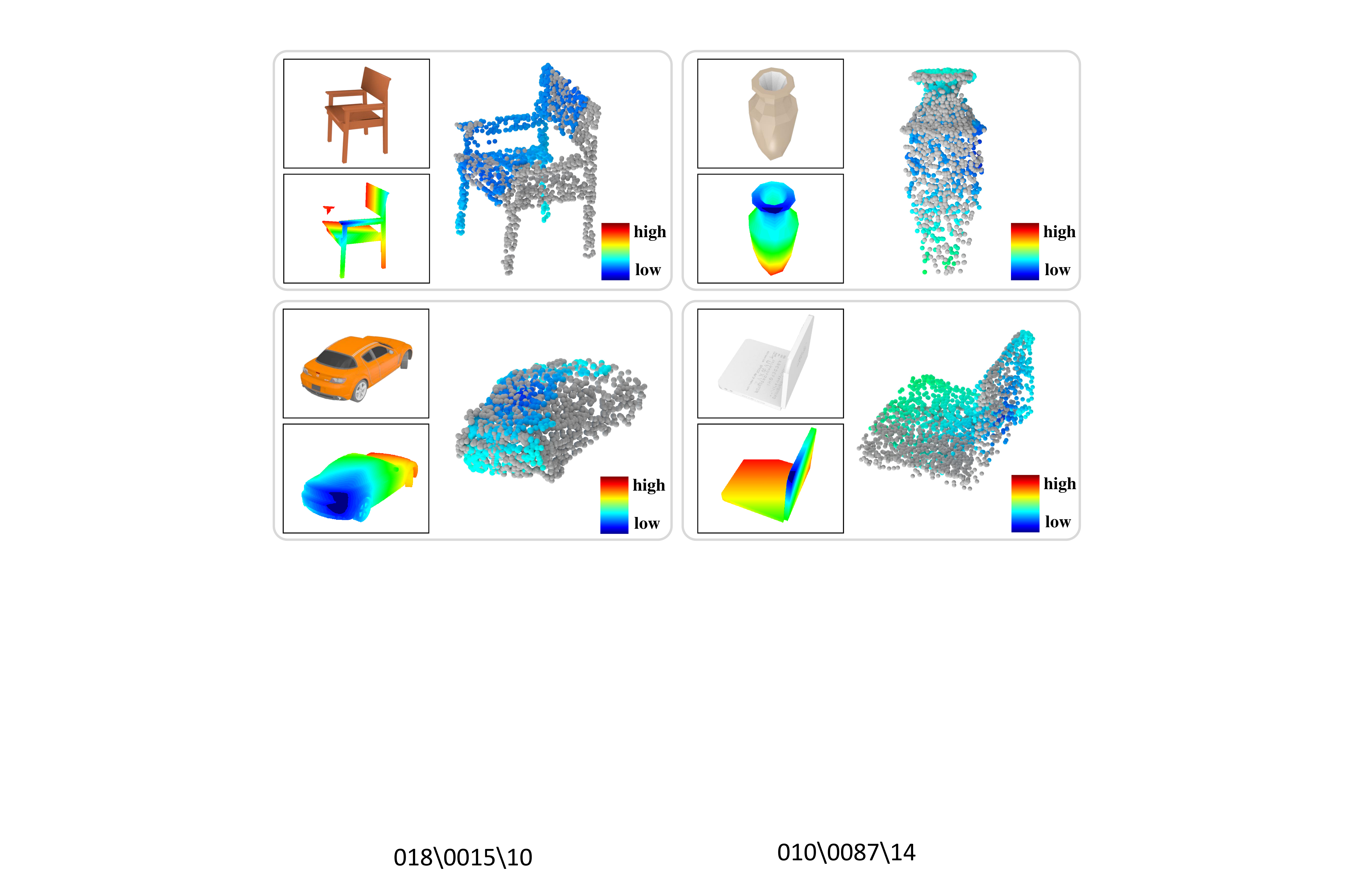}%,grid
   \end{overpic}
   \caption{\ys{Visualization of predicted counterparts. The grey points represent the input points. The colored points are the predicted counterparts, where color encodes the Euclidean distance error of the predictions.}}
   \label{fig:cpt_visual}
\end{figure}

\nc{
\subsection{Evaluation of counterpart prediction}
To evaluate the quality of the predicted counterparts, we compute and plot the distribution of the Euclidean distance of each predicted counterpart to its ground-truth counterpart, similar to~\cite{kim2011blended}.
Figure~\ref{fig:cpt_eval} shows the plots on the three ShapeNet subsets \ys{for reflectional symmetry}.
The x-axis of the plots represents a varying Euclidean distance (error) threshold. The y-axis shows the percentage of counterpart correspondences whose Euclidean distance are within the threshold.
Larger AUC (Area Under the Curve) represents better performance.
Compared to the baselines, our method is more accurate on the counterpart prediction task over all ShapeNet subsets.
\ys{Figure~\ref{fig:cpt_visual} shows a visualization of predicted counterparts.}
%More that 60% of the predicted points a counterpart error . This indicates the accurate counterpart prediction of our network.
}

%!TEX root = ../sceneparse.tex

%!TEX root = ../sceneparse.tex

\subsection{Prediction of discrete rotational symmetry}
\begin{figure}[t!]
	\begin{overpic}[width=1.0\linewidth,tics=10]{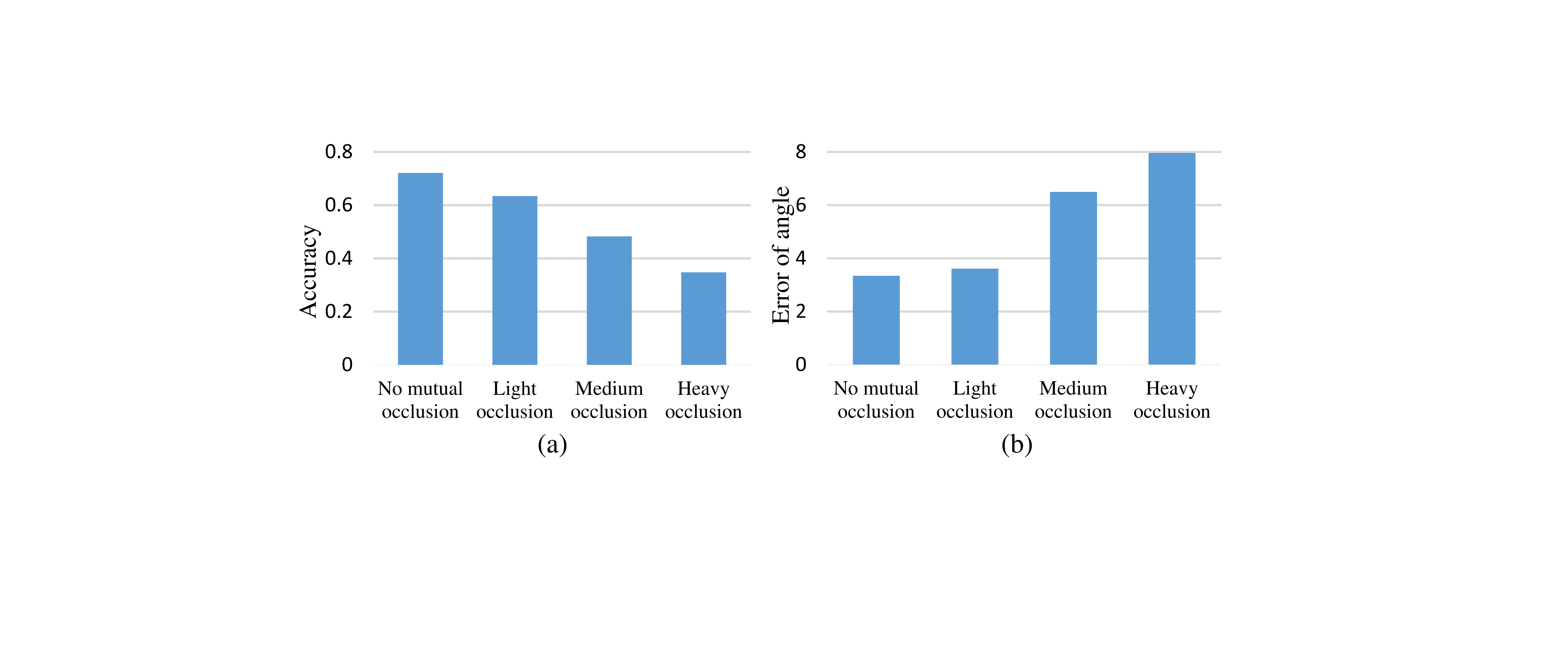}%,grid
   \end{overpic}
   \caption{\ys{Evaluation of discrete rotational symmetry prediction of our method. The plots show the classification accuracy (left) and the error of angle (right) on input with different occlusion ratios (the ratio of occluded area over the whole bounding box of the target object): no mutual occlusion, light (50-60\%) occlusion, medium (60-70\%) occlusion and heavy (70-80\%) occlusion. The error of angle is in degrees.}}
   \label{fig:discrete_rot}
\end{figure}

\ys{To evaluate the performance of our method on predicting discrete rotational symmetry, we create a dataset of shapes from ShapeNet, with ground-truth rotational symmetries of various orders.
We randomly add foreground occlusions to the rendered RGB-D images to verify the sensitivity to occlusion.
Details of the dataset can be found in the supplemental material.
We use two metrics to evaluate the results. First, since the order prediction is a classification task, classification accuracy is used. Second, the error of rotational angle is reported. Note that the error of rotational angle is computed as %$\vert\frac{2\pi}{r}-\frac{2\pi}{\hat{r}}\vert$.
$\vert2\pi/r-2\pi/\hat{r}\vert$, where $r$ and $\hat{r}$ are the predicted and the ground-truth order of the rotational symmetry, respectively.
Figure~\ref{fig:discrete_rot} shows the results.
We see that our proposed method can accurately predict the order of discrete rotational symmetries when the occlusion is not present or light. The performance drops as the occlusion increases.
}

%!TEX root = ../sceneparse.tex

\subsection{Runtime analysis}
Table \ref{tab:runtime} reports the timing of each component in our approach on a server with an Intel\textsuperscript{\textregistered} Xeon\textsuperscript{\textregistered} CPU E5-2678 v3 @ 2.50GHz $\times$ 48, 128GB RAM, and an Nvidia TITAN V graphics card.
Note that our method is dramatically faster compared to the state-of-the-art method in~\cite{ecins2018seeing} which takes about $15$ seconds to detect symmetries for an object.

\begin{table}
\caption{
Training and runtime performance statistics of each component of the proposed SymmetryNet.
}
\label{tab:runtime}
\vspace{-1ex}
\footnotesize\centering
\begin{tabular}{ccccc}
\toprule
 Dataset & Network train & Network inference & Aggregation & Verification \\
\midrule
ShapeNet & $64$ h & $50$ ms & $50$ ms & $40$ ms \\
%\midrule
YCB & $20$ h & $50$ ms & $50$ ms & $40$ ms \\
%\midrule
ScanNet & $26$ h & $50$ ms & $50$ ms & $40$ ms \\
\bottomrule
\end{tabular}
\end{table}

%!TEX root = ../sceneparse.tex

\begin{figure}[t]
	\begin{overpic}[width=1.0\linewidth,tics=10]{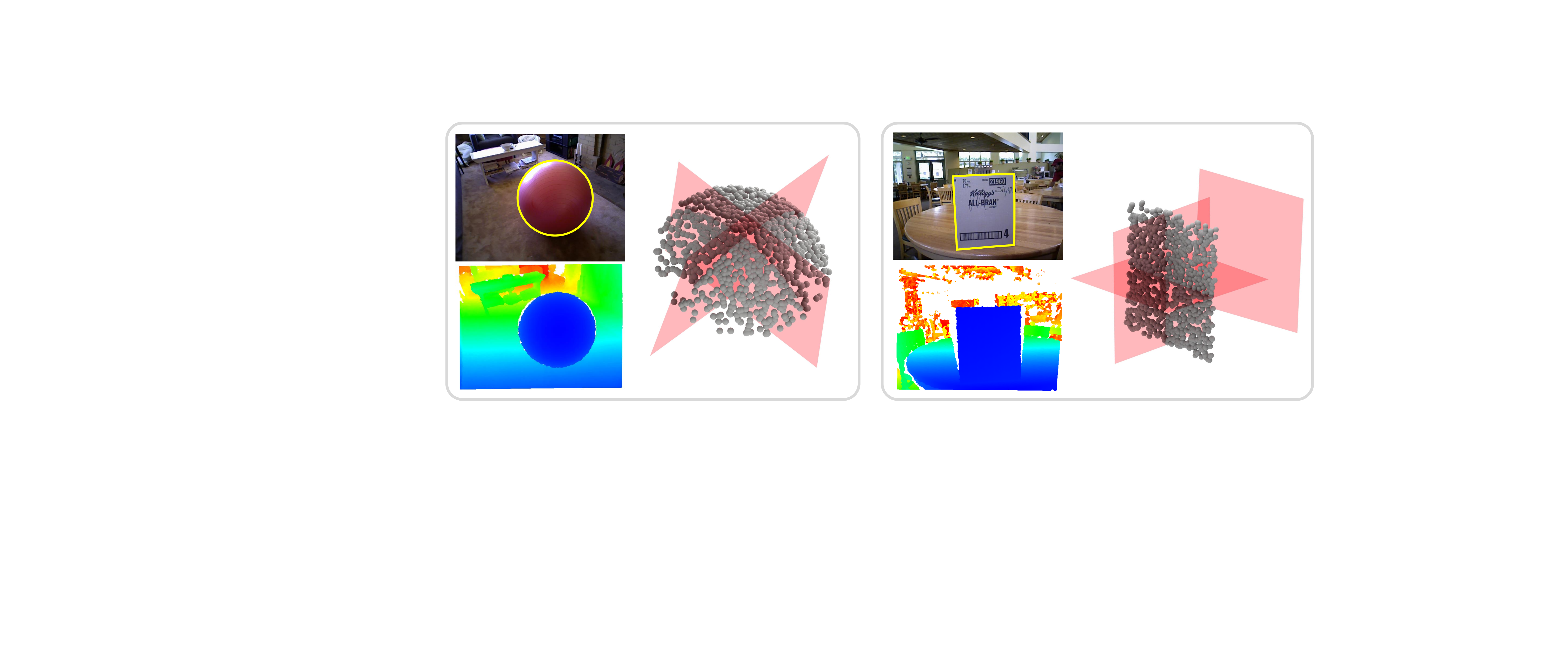}%,grid
   \end{overpic}
   \caption{\nc{Two failure cases. Left: Our method cannot deal with spherical symmetry characterized by the center of a sphere; It detects two reflectional symmetries wrongly. Right: When a cuboid object is viewed orthogonally from only one face, our method would fail to infer the reflectional symmetry in the depth direction.
   }}
   \label{fig:failure}
\end{figure} 

\nc{
\subsection{Failure cases}
\Fig{failure} shows two typical failure cases found in our experiments. The first case is that our method is unable to deal with other symmetry types than it was trained for. In the example, it detects two reflectional symmetries for a spherical symmetry. Another case is when a cuboid object is viewed orthogonally from only one face, our method would fail to infer the location of the reflection plane along the depth direction since the shape information is completely missing along that direction.
}

\subsection{Applications}
Various applications can potentially benefit from symmetry prediction. A straightforward application is symmetry-based object completion as commonly demonstrated in many symmetry detection works (e.g., ~\cite{bokeloh2009symmetry}). Here, we focus on a more unique application, i.e., how to apply our predicted symmetries to assist 6D object pose estimation from single-view RGB-D images.
%Here, we describe how to apply our predicted symmetries to two applications: object reconstruction and 6D pose estimation.

%It is straightforward to use our predicted symmetries to achieve object reconstruction with RGB-D images of a single object by augmenting the measured points with points transformed by the predicted symmetries.
%Note that although we have the predicted counterpart for each point in the symmetry prediction network, we choose to use the final predicted symmetry after the inference stage to ensure the accuracy and consistency of the predicted counterparts.
%We perform Poisson surface reconstruction~\cite{kazhdan2006poisson} on the completed point cloud to obtain a final reconstructed surface.
%Figure ? visualizes the predicted symmetry, predicted counterparts and the reconstructed surface of two objects.

\begin{figure}
	\begin{overpic}[width=1.0\linewidth,tics=10]{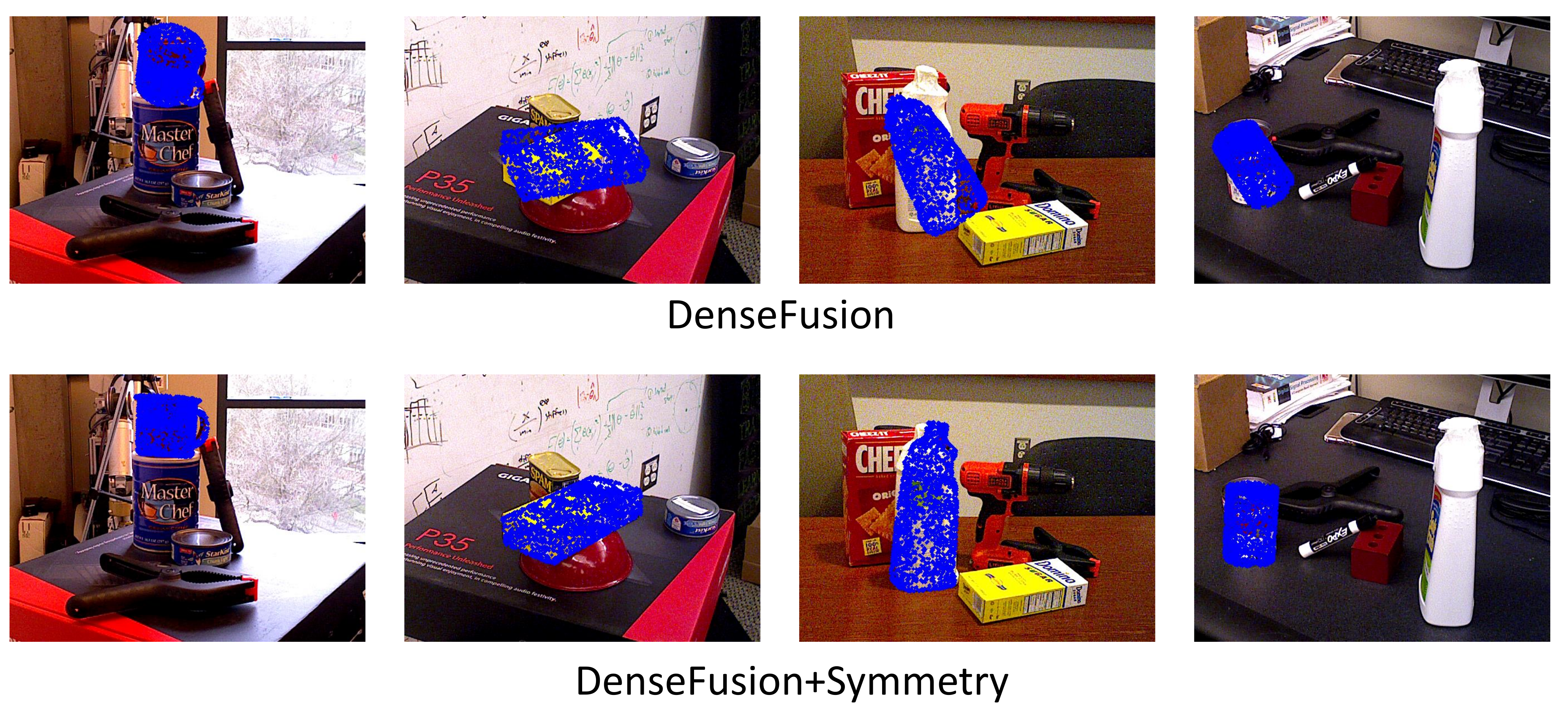}%,grid
   \end{overpic}
   \caption{By incorporating the feature of our predicted symmetries, DenseFusion+Symmetry achieves more accurate pose estimations, compared to the original DenseFusion approach~\cite{wang2019densefusion}.
   }
   \label{fig:app_6d}
\end{figure}

To demonstrate the effectiveness of the predicted symmetries on the improvement of 6D pose estimation, we combine the predicted symmetry information of our approach to the state-of-the-art 6D pose estimation approach DenseFusion~\cite{wang2019densefusion}. To be specific, we feed the parameters of predicted symmetries to DenseFusion as extra features. These features are then processed by an MLP and concatenated to the point cloud feature in DenseFusion. We train the network using the same data as described in~\cite{wang2019densefusion}.
%Details of the network are presented in appendix ?.
Figure~\ref{fig:app_6d} demonstrates how the predicted symmetries boost the performance of DenseFusion.

\begin{figure}[t!]
	\begin{overpic}[width=1.0\linewidth,tics=10]{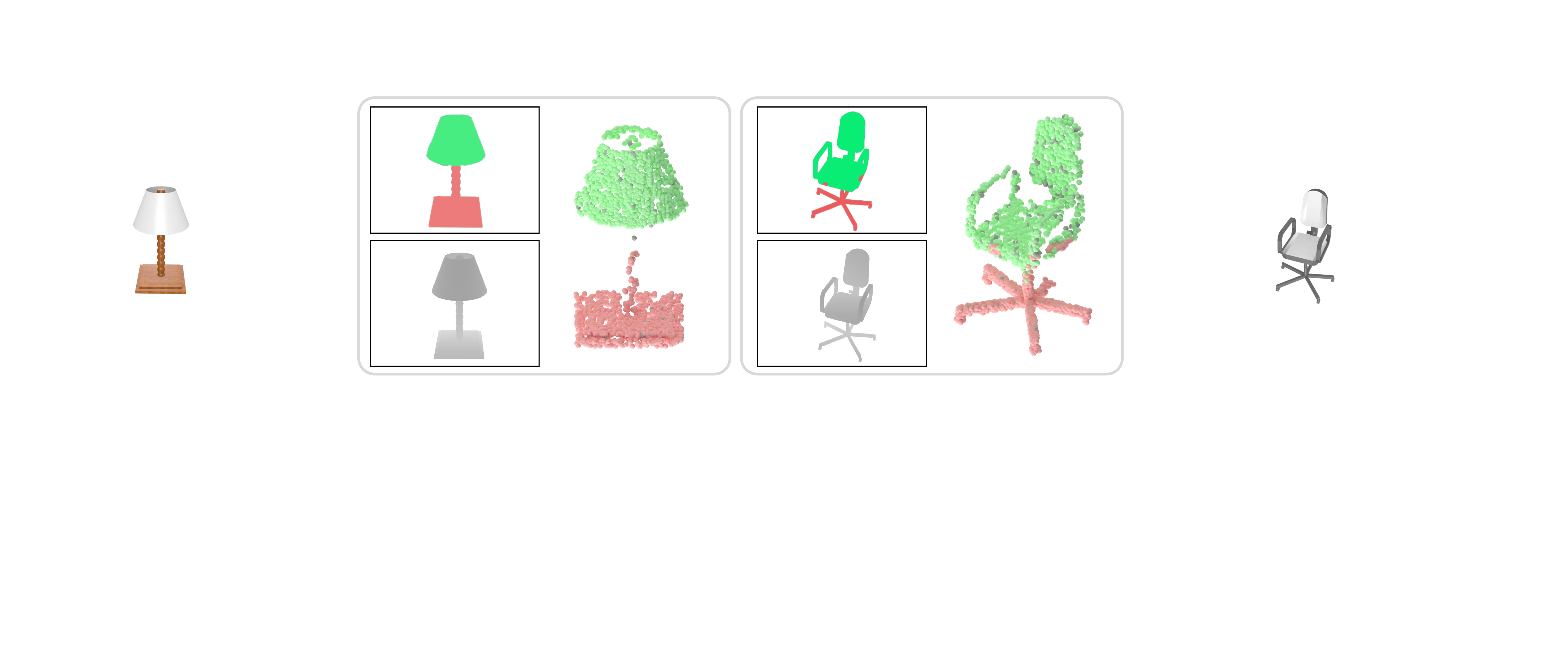}%,grid
   \end{overpic}
   \caption{\ys{Symmetry-induced RGB-D segmentation. The input RGB-D images are segmented through projecting the points segmented and labeled with different predicted symmetries. The segmentations are visualized by distinct colors both on the 3D point clouds and the input RGB images. Note that the outlier points are colored grey.}}
   \label{fig:segmentation}
\end{figure}

\ys{Another interesting application is symmetry-induced segmentation of RGB-D images. Thanks to our symmetry prediction, the input RGB-D images can be segmented into parts presenting different symmetries, although there was no such segmentation label for RGB-D images available during training.
This can be realized naturally by projecting the point-wise symmetry labels to the input RGB-D images.
%our method does not consider any segmentation label during training, it is still able to segment the input point cloud with the help of the point-wise predicted symmetries.
Figure~\ref{fig:segmentation} demonstrates two examples of this application.}
%!TEX root = ../sceneparse.tex

%!TEX root = ../sceneparse.tex

%!TEX root = ../sceneparse.tex

%!TEX root = sceneparse.tex

\section{Discussion and conclusions}
\label{sec:future}
We have proposed a novel problem of detecting 3D symmetries from single-view RGB-D images. Due to partial observation and object occlusion, the problem is challenging, to the point of being beyond the reach of purely geometric detection methods. Instead, we have proposed an end-to-end deep neural network that predicts both reflectional and rotational symmetries for 3D objects based on a single RGB-D image. Several dedicated designs make our method general and robust. \emph{First}, our network is trained on multiple tightly coupled tasks to achieve outstanding generalizability for both types of symmetries. \emph{Second}, we devise an optimal assignment module in our network to enable it to output an arbitrary number of symmetries.

Our current method has certain limitations, which we believe will inspire future research:
\begin{itemize}
  \item Our method relies on a good object-level segmentation. If the segmentation mask of an object of interest contains other objects, the symmetry detection will be affected. Although there have been many powerful deep models trained for RGB-D segmentation, it would still be interesting to integrate object detection/segmentation and symmetry detection into one unified deep learning model.
  \item Our current network can only deal with reflectional and rotational symmetries. Extending it to other types of symmetry should not be difficult, although it may make the network harder to train. In general, finding a suited parameterization/representation of symmetry for end-to-end learning is a fundamental and interesting future direction to pursue.
  \item Our method cannot handle hierarchical (nested) symmetries such as those considered in~\cite{wang2011symmetry}. We expect that recursive neural networks (RvNN) could be utilized for this case, following the series of works on using RvNNs for 3D structure encoding/decoding~\cite{li2017grass,yu2019partnet}.
  \item Our network relies on strong supervision. Annotating symmetries for RGB-D data is a non-trivial endeavor. Therefore, it would be interesting to look into unsupervised or self-supervised approaches to symmetry detection, through exploiting rich geometric constraints.
\end{itemize} 

\begin{acks}
We thank the anonymous reviewers for their valuable comments. We are grateful to Yao Duan, Dengsheng Chen and Yuqing Lan for their discussion in data preparation. This work was supported in part by National Key Research and Development Program of China (2018AAA0102200, 2018YFB1305105), NSFC (61825305, 61751311, 61572507, 61532003, 61622212), Youth Innovation Project of College of Intelligence Science and Technology, NUDT (2020006, 2020008) and NSF grant IIS-1815070.
\end{acks}

{\small
\bibliographystyle{plainnat}
\setlength{\bibsep}{0pt}
\bibliographystyle{abbrvnat}
\bibliography{symmetry_prediction}
}

\end{document}